%% file: main.tex
\documentclass{article}
\usepackage{iclr2027_conference,times} % ICLR 2027 template; submission is anonymous unless the following is uncommented
\iclrfinalcopy

\usepackage[utf8]{inputenc}
\usepackage[T1]{fontenc}
\usepackage{url}
\usepackage{booktabs}
\usepackage{amsmath,amssymb}
\usepackage{amsthm}

\newtheorem{lemma}{Lemma}
\newtheorem{corollary}{Corollary}
\newtheorem{proposition}{Proposition}

\theoremstyle{remark}

\usepackage{graphicx}
\usepackage{wrapfig}
\usepackage{multirow}
\usepackage{xcolor}
\usepackage{xspace}
\usepackage{microtype}
\usepackage{enumitem}
\usepackage{caption}
\usepackage{placeins}
\graphicspath{{figs/}{.}}
\definecolor{hypercolor}{RGB}{140, 157, 175}
\usepackage[pagebackref,breaklinks,colorlinks,allcolors=hypercolor]{hyperref}
\usepackage[capitalize,noabbrev]{cleveref}

\usepackage{framed}
\definecolor{shadecolor}{gray}{0.94}

\newcommand{\fp}{\textsc{FP}\xspace}
\newcommand{\ptq}{\textsc{PTQ}\xspace}

\newcommand{\cset}{\ensuremath{\mathcal{C}_\varepsilon}\xspace}
\newcommand{\gapk}[1]{\ensuremath{g_{#1}}}
\newcommand{\qmargin}{\ensuremath{g_{\mathrm{q}}}\xspace}
\newcommand{\rtn}{\textsc{RTN}\xspace}
\newcommand{\gptq}{\textsc{GPTQ}\xspace}
\newcommand{\awq}{\textsc{AWQ}\xspace}

\title{The Undetected Damage of Quantization on Retrieval and How to Fix It}

\author{%
  \hfill Luca Zhou$^{1}$ \hfill Alessandro Zirilli$^{1}$ \hfill Daniele Solombrino$^{1}$ \hfill Roberto Dessì $^{3}$ \hfill Emanuele Rodolà$^{1,2}$ \hfill \\[1.5ex]
  {\small $^{1}$Sapienza University of Rome \quad $^{2}$Paradigma \quad $^{3}$Not Diamond}
}

\begin{document}
\maketitle
\lhead{Preprint}
{\let\thefootnote\relax\footnotetext{Correspondence: \texttt{luca.zhou@uniroma1.it}. Code: \url{https://github.com/LuckerZOfficiaL/Undetected-Damage-of-Quantization}}}

\begin{abstract}
We show that a quantized model that keeps its classification accuracy still changes $14$ to $46\%$ of its top-1 retrieval results, and that aggregate ranking metrics reveal only part of this damage. We tie this failure to the gap between the two highest model scores and use that gap to decide when to trust a quantized answer and where additional precision should be spent. We show that the top-1 result is guaranteed to survive quantization only when this gap exceeds twice the largest rounding error. In classification, the scores are logits, and training compares the correct class against every other class, which encourages this gap. In retrieval, the scores are query-document similarities, and training compares each positive only against sampled negatives, so nothing separates the top-1 item from the second. This gap can be measured without labels. Before deployment, it predicts which models will break under quantization, and at deployment time it tells, per input, whether the quantized answer still matches the full-precision answer. Most classification inputs have a gap wide enough to trust the quantized answer, but few retrieval queries do. That gap motivates a different fix in each task. In retrieval, spending extra bit-width on the layers whose quantization moves the gap most recovers up to three-quarters of an extra bit's benefit for half its cost. In classification, routing the few low-gap inputs to full precision recovers most of the lost accuracy at a fraction of the cost.

%Quantization rounds weights to a lower bit-width, and it is usually judged by how much accuracy it loses. This criterion is incomplete. A quantized model that keeps its classification accuracy still changes $14$--$46\%$ of its top-1 retrieval results, and the aggregate ranking metrics show only part of that change. The top-1 result is guaranteed to survive quantization only when the gap between the two highest scores exceeds twice the largest rounding error. The scores are logits in classification and query-document similarities in retrieval. In classification, training compares the correct class against every other class, which opens this gap. In retrieval, training compares each positive only against sampled negatives, so nothing separates the top-1 item from the second. This gap can be measured without labels. Before deployment, it predicts which models will break under quantization, and at deployment time it tells, per query, whether the quantized answer still matches that of the full-precision model. The check passes for most classification inputs but only for a minority of retrieval queries. That gap motivates a different fix in each task. In retrieval, spending extra bit-width on the layers whose quantization moves the gap most recovers up to three-quarters of an extra bit's benefit for half its cost. In classification, routing the few low-gap inputs to full precision recovers most of the lost accuracy at a fraction of the cost.
\end{abstract}

\begin{wrapfigure}[20]{r}{0.48\linewidth}
\vspace{-1.9em}
\centering
\includegraphics[width=\linewidth]{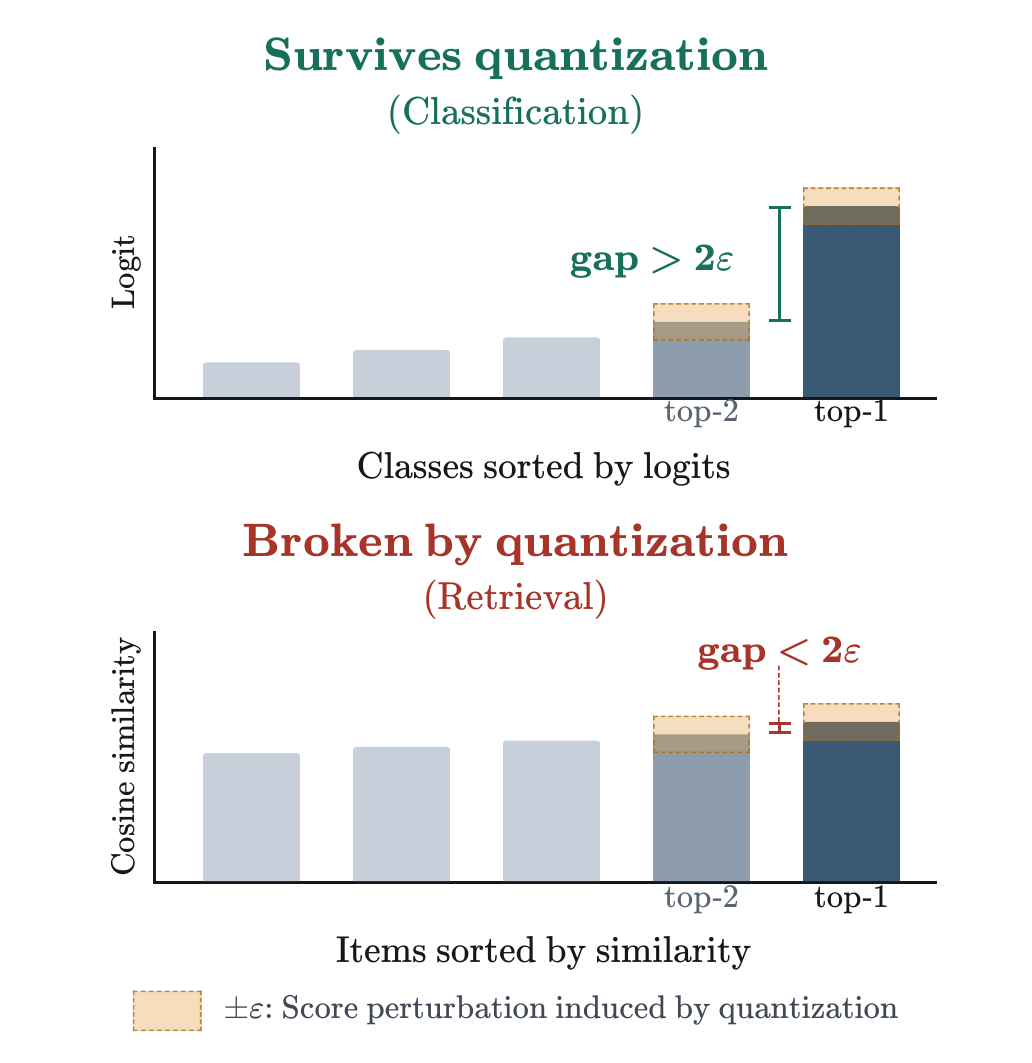}
\vspace{-2em}
\caption{\textbf{What quantization preserves and breaks.} Every score moves by at most $\varepsilon$, so a top-1 result survives whenever it leads the runner-up by $2\varepsilon$.}
\label{fig:teaser}
\end{wrapfigure}

% =============================================================================
\section{Introduction}

Quantization compresses a neural network by rounding its weights to a lower bit-width. Post-training quantization (\ptq{})~\citep{gholami2022survey} is often evaluated by how much accuracy the model loses relative to full precision. A 4-bit model that nearly matches its full-precision checkpoint in accuracy is declared a success, and for classification that intuition is right. But accuracy measures one thing only: whether the top-1 prediction changes, yet most deployed models are not classifiers. Retrieval, retrieval-augmented generation, recommendation, and reranking all
depend on an \emph{ordering}. For them, we find that a quantized model that preserves full-precision classification accuracy can still rank items very differently, including at the top-1. This problem goes undetected by metrics of quantization performance.

The effect is large and consistent. Across three backbones and two modalities, 4-bit round-to-nearest weight quantization (W4) changes the top-1 prediction of a fine-tuned checkpoint for only $3$--$6$\% of inputs when it is used as a classifier, but for $33$--$46$\% of queries when the same model is used for retrieval. The aggregate metrics understate this issue. On CLIP~\citep{radford2021clip} text-to-image retrieval, Recall@1 falls by only $1.3$\% while about one in ten correctly answered queries loses its image. On Qwen3-Embedding-8B~\citep{zhang2025qwen3}, nDCG@10 drops by only $3.1$\% while one in six queries with a relevant top result loses it.

The reason is which candidates each task family trains against. A ranking is stable under quantization only where the gap between two scores exceeds twice the worst-case rounding error, whether the scores are logits in classification or query-document similarities in retrieval. Classification trains against a fixed, complete set of classes, so every class is pushed below the correct one at every step, and the top-1 logit ends up far from the runner-up. Retrieval trains against a sampled subset of documents, so a positive is separated only from the negatives it was paired with, and nothing pushes the top document away from the second (Figure~\ref{fig:teaser}). To make this exact, let $\varepsilon$ be the largest amount by which quantization perturbs any score. Since the top-1 can fall by $\varepsilon$ while a challenger rises by $\varepsilon$, a candidate can overtake the top-1 only if it falls behind by less than $2\varepsilon$. When no candidate is that close, the top-1 cannot change, on the other hand, when some are, the number of them predicts how likely the result changes. For most classification inputs, no candidate is that close, while for nearly every retrieval query, several are. This is why the same quantization leaves classification accuracy intact but breaks retrieval rankings. The same gap accepts or rejects individual answers at deployment time, without labels. We compare the quantized model's own top-1/top-2 gap against a single threshold, calibrated in advance on unlabeled data. The comparison tells us, for each input, whether the quantized answer is likely to match that of full precision model. Most classification inputs pass this check, but few retrieval queries do, given their typically small gaps. The damage is not an artifact of naive rounding, as quantizers built to minimize rounding error, such as \gptq{}~\citep{frantar2023gptq}, \awq{}~\citep{lin2024awq}, HQQ~\citep{badri2023hqq} and AdaRound~\citep{nagel2020adaround}, still show the same failure.

Since nearly every retrieval query is at risk, the intervention must apply before deployment. In a mixed-precision setting, we allocate extra bit-width to the layers whose quantization most perturbs the gap between the top two items. In classification, only a small fraction of inputs is at risk, so the intervention can be adaptive with a first quantized forward pass estimating the gap, and one threshold on it routes the at-risk inputs to the full-precision model. Contributions are summarized below:

\begin{enumerate}[leftmargin=1.4em,itemsep=2pt]
  \item \textbf{Quantization breaks retrieval when accuracy holds.} When we hold a checkpoint fixed and change only the task, the top-1 retrieval result at W4 changes $5$--$23\times$ more often than the classification answer (Tables~\ref{tab:fragility} and~\ref{tab:regime}). Standard retrieval metrics fail to detect this change. The damage is not tied to the quantizer, since stronger ones do not solve the issue.
  \item \textbf{One label-free quantity explains and predicts it.} The gap between the top two scores, measured against the perturbation, explains why retrieval breaks where classification does not, predicts the damage across models, quantizers, and corpora, and tests individual inputs without labels.
  \item \textbf{Two fixes from one quantity.} On retrieval, allocating bit-width by gap sensitivity recovers three-fifths to three-quarters of an extra bit's benefit for half its cost and outperforms every criterion under every quantizer we test. On classification, the same gap, thresholded once, routes the inputs at risk to full precision, recovering most of the lost accuracy at a fraction of the full-precision cost.
\end{enumerate}

% =============================================================================
\section{Related work}

\textbf{Post-training quantization.} Post-training methods take a trained model and fit the
quantization to a small unlabeled calibration set, choosing scales, rounding, or a rotation of
the weights so that each quantized layer reproduces the outputs of the original one:
\gptq{}~\citep{frantar2023gptq}, \awq{}~\citep{lin2024awq}, SmoothQuant~\citep{xiao2023smoothquant},
QuaRot and SpinQuant~\citep{ashkboos2024quarot,liu2024spinquant}, and for vision transformers
PTQ4ViT, RepQ-ViT and AdaLog~\citep{ptq4vit,li2023repqvit,adalog}, surveyed
by~\citet{gholami2022survey}.

\textbf{Ranking under compression.} On mixture-of-experts routing, rank-preservation losses keep a quantized router choosing the experts that full precision would choose. A separate analysis notes that a small perturbation from quantizing the KV cache can change which experts top-$k$ routing selects, but it does not measure how often~\citep{vsraq2026,routeflip2026}. On classification, perturbation bounds have been stratified by the top-2 logit gap for the true-label margin~\citep{boundaryaware2026}.
On ranking under uncertainty, when scores are intervals, two items can be compared only if their intervals do not overlap~\citep{soliman2009ranking}. When every interval has the same width, the relation is a semiorder~\citep{luce1956semiorders,scott1958foundational,fishburn1970intransitive}, and the rankings a bounded perturbation can reach are its linear extensions~\citep{szpilrajn1930extension}. We restate the top-1 case in the notation the rest of the paper uses. What is new here is the measurement, not the condition.
On embedding compression, the stored document vectors are quantized~\citep{jegou2011searching,guo2020anisotropic}
or the query encoder is trained jointly with the index~\citep{zhan2021jpq}, and reconstruction
error is known to be the wrong objective for ranking~\citep{guo2020anisotropic}. We differ in
three ways. We quantize the encoder itself rather than the index, we read one model both as a
retriever and as a classifier, and we test individual inputs at deployment time without labels.

\textbf{Mixed-precision allocation.} Per-layer bit-width assignment is established for classification~\citep{dong2019hawq,dong2020hawqv2,yao2021hawqv3}, with ZeroQ replacing curvature by a KL divergence to the full-precision outputs~\citep{cai2020zeroq} and mixed precision reaching the KV cache of long-context inference~\citep{tao2025moqae}. All target output errors. None measures what a layer's quantization does to the ranking in retrieval.

\textbf{Selective prediction and routing.} Selective prediction sends inputs a model is unsure about to a stronger one, scored by maximum softmax probability~\citep{hendrycks2017baseline,geifman2017selective} or by the margin between the best and second-best class, first used as an uncertainty score for active learning~\citep{joshi2009multi}. Cascades, early exits, and speculative decoding with quantized drafts rely on the same mechanism~\citep{teerapittayanon2016branchynet,huang2018msdnet,kolawole2024revisiting,georganas2025mlspecqdmultilevelspeculativedecoding,zhao2025qspec}. Compression changes predictions while accuracy holds~\citep{hooker2019compressed,hooker2020characterising,dutta2024accuracy}, and among single scores the margin is the best predictor of which predictions change~\citep{qiu2022characterizing}.

% =============================================================================
\section{When a ranking can change}\label{sec:theory}

The rounding error introduced by quantizing the weights propagates through the forward pass and displaces every score by a bounded amount. Although we measure the top-1 throughout, we analyze entire rankings because a retrieval top-1 is the first position of a ranking over thousands of candidates, and what displaces it is often a candidate from deep in the list rather than the runner-up. Formally, let $z \in \mathbb{R}^N$ be the \emph{score vector} over $N$ candidates, with $z_i$ the score of candidate $i$. Throughout, a \emph{score} is whatever quantity the system ranks by, a class logit in classification and a query-candidate similarity in retrieval. The analysis below applies to both cases. A \emph{permutation} $\sigma = (c_1, \dots, c_N)$ lists the candidates, so that $c_k$ is the candidate in $k$-th position. The \emph{ranking} of $z$, denoted $\pi(z)$, is the permutation that sorts $z$, \emph{i.e.}\ the one for which $z_{c_1} > \dots > z_{c_N}$, and it is undefined when $z$ has ties. Given $\varepsilon > 0$, a \emph{perturbation} is any $\delta \in \mathbb{R}^N$ with $\lVert \delta \rVert_\infty \le \varepsilon$. 

A permutation $\sigma$ is \emph{$\varepsilon$-reachable} from $z$ if some perturbation $\delta$ has $\pi(z + \delta) = \sigma$. We write $\mathcal{R}_\varepsilon(z)$ for the set of reachable permutations. It always contains $\pi(z)$, via $\delta = 0$. We model quantization as replacing $z$ with $z + \delta$ at the output, where $\varepsilon$ depends on the bit-width.

Define the $k$-th \emph{gap} as $\gapk{k}(z) = z_{c_1} - z_{c_k}$, the difference between the first score and the $k$-th, so that $\gapk{1} \equiv 0$ and $\gapk{k}$ is non-decreasing in $k$. Since a perturbation can change one score by at most $\pm\varepsilon$, the $k$-th candidate can overtake the winner only if $\gapk{k}(z) < 2\varepsilon$. 
We therefore call $\cset(z) = \{\, k \ge 2 : \gapk{k}(z) < 2\varepsilon \,\}$ the \emph{contender set}, \emph{i.e.}\ the positions that a perturbation could possibly push past the top-1, and $\cset(z) = \emptyset$ means the top-1 has no contender and cannot be surpassed.

Given an input $x$, we write $z(x)$, $\gapk{k}(x)$ and $\cset(x)$ for the scores, gaps and contenders it determines. These are obtained by running both the full-precision (\fp{}) model ($\cdot^{\fp}$) and its quantized counterpart ($\cdot^{\ptq}$) on $x$ and measuring, not assuming, the perturbation $\delta = z^{\ptq} - z^{\fp}$, from which we take $\varepsilon = \lVert \delta \rVert_\infty$. We call $\gapk{2}(x)/2\varepsilon$ the \emph{separation ratio}. It is at least one exactly when $\cset(x) = \emptyset$, so we call one the \emph{stability threshold}. Across inputs, a median separation ratio above one means the typical input has a stable top-1. Every diagnostic below computes it from the \fp{} gap and the perturbation.

\begin{lemma}[Top-1 and top-$k$ stability]\label{lem:cuts}
Write $\pi(z) = (c_1, \dots, c_N)$. For each $k$ from $1$ to $N-1$, every permutation $\sigma \in \mathcal{R}_\varepsilon(z)$ has the same first $k$ elements as $\pi(z)$, as a set, iff $z_{c_k} - z_{c_{k+1}} \ge 2\varepsilon$. For $k = 1$ this says the top-1 survives every perturbation iff $\gapk{2}(z) \ge 2\varepsilon$, that is, iff $\cset(z) = \emptyset$.
\end{lemma}

Each score moves by at most $\varepsilon$, so a candidate can overtake another only if that other started at most $2\varepsilon$ ahead of it. When the gap is smaller, the proof of Lemma~\ref{lem:cuts} in Appendix~\ref{app:proofs} constructs a perturbation that swaps the two. Both conditions are known. They restate, in the notation of this paper, the classical case in which each score is known only up to an interval of fixed width~\citep{luce1956semiorders,scott1958foundational,fishburn1970intransitive,szpilrajn1930extension}. Appendix~\ref{app:theory-detail} states the general characterization of reachable rankings (Proposition~\ref{prop:reachable}) and a closed-form flip probability under a uniform perturbation model (Proposition~\ref{prop:pflip}). The top-$k$ condition would be the natural guarantee for ranked retrieval, but the gaps below the top-1 are an order of magnitude too small for it to be usable in practice (Table~\ref{tab:topk}), so everything we propose rests on the top-1 condition. Being within $2\varepsilon$ is not transitive, so a chain of near-ties lets a candidate far below the top-1 overtake it, which is where a large share of retrieval flips come from. The first gap is still enough to diagnose the risk.

\begin{corollary}[Contender locality]\label{cor:local}
Let $a$ be the top-1 of $z$. Candidate $j$ overtakes $a$ under perturbation $\delta$ iff $\delta_j - \delta_a > z_a - z_j$, so whether the top-1 survives depends on $\delta$ only through the differences $\{\delta_j - \delta_a\}$ and not on $\lVert \delta \rVert_\infty$ over all $N$ coordinates. Consequently, writing $a'$ for the top-1 of $z + \delta$ and $S = \max_j (\delta_{a'} - \delta_j)$, a change of top-1 implies $\gapk{2}(z + \delta) < S$, and $S \le 2\lVert \delta \rVert_\infty$.
\end{corollary}

%\az{Only differences matter: a perturbation that moves every score by the same amount leaves every ranking untouched, however large the perturbation happens to be. This sentence is either truncated or very odd to me} 
Only differences matter: a perturbation that moves every score by the same amount leaves every ranking untouched, however large the perturbation happens to be. The global bound $2\varepsilon$ ignores this since it also counts a shift that the winner and a challenger share, which cancels in their difference. That is why the stability check below calibrates $S$ rather than $2\varepsilon$.

\textbf{From condition to stability check.}\label{par:certificate}
The top-1 condition of Lemma~\ref{lem:cuts}, $\gapk{2}(z^{\fp}) \ge 2\varepsilon$, is stated with the full-precision scores and the size of the perturbation, and at deployment time only the quantized model runs, so neither is available. The condition can, however, be rewritten in terms of the quantized scores alone. Write $c'_1$ for the candidate the quantized model ranks first, $\gapk{2}(z^{\ptq}(x))$ for its top-1/top-2 gap, and $S(x) = \max_j (\delta_{c'_1} - \delta_j)$ for the largest amount by which $\delta$ moves a score difference involving $c'_1$. Whenever the top-1 changes, $\gapk{2}(z^{\ptq}(x)) < S(x)$, and $S(x)$ never exceeds $2\varepsilon$ (Corollary~\ref{cor:local}). So if the quantized gap is at least $S(x)$, the top-1 did not change, and unlike $\varepsilon$, $S$ can be calibrated once and stored. On $n$ unlabeled calibration inputs, run both models, compute $S$ on each, and let $\hat\tau$ be the $\lceil (n+1)(1-\alpha) \rceil$-th largest value, the split-conformal quantile. At deployment time, accept a test input, i.e., trust its quantized top-1 as the full-precision answer, iff $\gapk{2}(z^{\ptq}(x)) \ge \hat\tau$. When the calibration and test inputs come from the same distribution, split-conformal calibration~\citep{vovk2005algorithmic,lei2018distfree} gives
\[
  \mathbb{P}\big(\text{accepted and top-1 changed}\big) \;\le\; \alpha ,
\]
with no labels at test time. Calibrating $2\varepsilon$ instead of $S$ is also valid but accepts far fewer inputs, and no
retrieval query at all (Appendix~\ref{app:certificate}). The calibration pass costs nothing extra, since a quantization pipeline already runs the \fp{} model on unlabeled data to calibrate quantizers such as \gptq{}.

% =============================================================================
\section{Setup}\label{sec:setup}\label{sec:setup-sub}

\textbf{Models.} We study three fine-tuned backbones across two modalities: ViT-B/16 and ViT-L/16~\citep{dosovitskiy2021image} for images, and Qwen3-Embedding-0.6B~\citep{zhang2025qwen3} for text, giving 53 fine-tuned checkpoints in total. Later sections add five more text embedders, three encoder-based~\citep{xiao2024cpack,wang2022e5,li2023gte} and Qwen3-Embedding-4B and 8B, together with \textsc{clip}~\citep{radford2021clip}, a cross-encoder reranker~\citep{nguyen2016msmarco}, and a Qwen3-8B generator for the retrieval-augmented pipeline. Quantization is done on weights at 4 and 3 bits (W4 and W3), with activations kept at full precision except in a specific experiment. 
Every linear layer of every model is quantized. Appendix~\ref{app:setup-detail} details fine-tuning and the remaining settings.

\textbf{Tasks.} Classification and retrieval each run in both modalities, and a fifth setting is a deployed cross-modal system. In image classification, each ViT is fine-tuned separately on 21 image classification tasks. In text classification, Qwen3-Emb-0.6B is fine-tuned on 11 tasks from \textsc{mteb}~\citep{muennighoff2023mteb}. In image retrieval, the same fine-tuned ViTs rank each task's test split by similarity to a query image drawn from that split. This compares classification and retrieval on the same model. In text retrieval, the Qwen3 embedders rank four standard \textsc{beir} corpora~\citep{thakur2021beir}, NFCorpus~\citep{boteva2016nfcorpus}, SciFact~\citep{wadden2020scifact}, SCIDOCS~\citep{cohan2020specter} and FiQA~\citep{maia2018fiqa}. In text-to-image retrieval, \textsc{clip} ranks $4{,}000$ Flickr30k images~\citep{young2014flickr30k} by similarity to a caption, with no fine-tuning. Section~\ref{sec:systems} adds two systems built on retrieval: a cross-encoder reranker on the same \textsc{beir} corpora and a retrieval-augmented generation pipeline on Natural Questions~\citep{kwiatkowski2019natural}.

\textbf{Evaluation.} We report the top-1 result throughout (the predicted class in classification, the first-ranked item in retrieval) and measure how often it differs from the \fp{} model. This needs no relevance labels, since it compares the two models rather than either one against ground truth. Six procedures need calibration data, and none takes labels. The allocation criteria use $128$ calibration queries by default, and the data-dependent quantizers use $512$ documents. Calibration and evaluation are disjoint throughout, with one exception described in the calibration protocol in Appendix~\ref{app:setup-detail}.

% =============================================================================
\section{What survives quantization}\label{sec:fragility}

\begin{wraptable}[11]{r}{0.53\linewidth}
\vspace{-1.2em}
\centering
\footnotesize
\caption{\textbf{Same backbone, two tasks.} ViT rows use one checkpoint for both tasks, the first Qwen3 row uses a fine-tuned classifier against the pretrained embedder, the last row uses one checkpoint, classified with a linear probe and retrieved by cosine self-retrieval. Top-1 change rate at W4 group\_128 under \rtn{}, mean $\pm$ standard error over $n$ tasks.}
\label{tab:fragility}
\resizebox{\linewidth}{!}{\input{tables/new_fragility}}
\end{wraptable}

The same fine-tuned model, under the same quantizer, at the same bit-width, changes its top-1 answer for a few percent of inputs when asked to classify and for over forty percent of queries when asked to retrieve (Table~\ref{tab:fragility}). The Qwen3 rows show the same asymmetry in another model family, first with a fine-tuned classifier against the pretrained embedder, then with one checkpoint read both ways. The retrieval top-1 changes $4.8$--$14.3\times$ more often than the classification top-1. The perturbation is similar in both cases, and what changes is the gap it has to cross. Classification trains against every class and retrieval against a sampled subset, so a positive is separated only from the negatives it was paired with, and the second place is a comparison training never made.

\begin{wraptable}[14]{r}{0.58\linewidth}
\centering
\scriptsize
\vspace{-1.7em}
\caption{\textbf{A mean quality metric understates per-query damage.} \emph{NDCG loss} is the relative change in mean nDCG@10. \emph{Gold lost} is the fraction of queries for which a document that was both relevant and in the \fp{} top-$k$ falls out of the \ptq{} top-$k$. Round-to-nearest, group\_128, mean over four \textsc{beir} corpora.}
\label{tab:gold}
\input{tables/new_gold}
\end{wraptable}

Most changed results are worse. Judged against the \textsc{beir} relevance labels over six embedders, four corpora and six quantizers, $66$\% of the flips whose \fp{} top-1 was a relevant document replace it with one that is not, and $77$\% do at three bits (Table~\ref{tab:harm}). A changed top-1 has two outcomes: either the new item is still relevant or it is not. So a $34$\% change rate does not by itself imply a loss of quality. On \textsc{clip}, where every query has one correct image, $26$--$30$\% of all changed results lose it and $13$--$23$\% gain it (Appendix~\ref{app:clip}, \emph{What a changed result costs}). The flips that keep a relevant item on top are not harmless either. A retriever is one component of a larger system, so a different document yields a different generated answer (\S\ref{sec:systems}), and two servers running the same model at different precisions answer the same query differently. The standard metric reveals only a fraction of this damage, harmful or not: nDCG@10 averages over ten positions, and the item that lost first place usually stays inside that window. Across three model families at W4, nDCG@10 falls by $2.7$--$7.6$\% while the relevant document leaves the top-1 for $16$--$22$\% of the queries that had one, and it leaves the top-10 just as often (Table~\ref{tab:gold}). Accuracy also tells nothing about the rest of the ranking. On classifiers whose top-1 accuracy is preserved, the exact top-5 set survives for only $28$-$47$\% of inputs (Table~\ref{tab:topk}).

% =============================================================================

\subsection{Nothing else explains the asymmetry}\label{sec:controls}

Nothing but the tasks explain why retrieval breaks more than classification. The asymmetry survives when everything else is held fixed. The last row of Table~\ref{tab:fragility} embeds one dataset once with one encoder and reads the same vectors both ways, through a frozen linear probe and by cosine nearest neighbor, and at W4 retrieval still changes $3.3$--$6.8\times$ more of its top-1 answers on each dataset (Table~\ref{tab:contrast}). Accuracy loss does not predict it. Within a quantization setting, the accuracy a configuration loses correlates with the fraction of top-$5$ sets it changes at only $+0.327$, and in the deployed W4-group\_128 setting the correlation is undetectable on either vision backbone (Table~\ref{tab:accrank}). The number of candidates does not predict it either. Growing the corpus by up to $58\times$ leaves the text change rate within three points and raises the vision rates sharply, and in both cases the top-1 change follows the separation ratio, not the candidate count (Table~\ref{tab:corpus}).

% =============================================================================

\subsection{The contender set is the mechanism}\label{sec:mechanism}

\begin{figure}[t]
\centering
\includegraphics[width=0.9\linewidth]{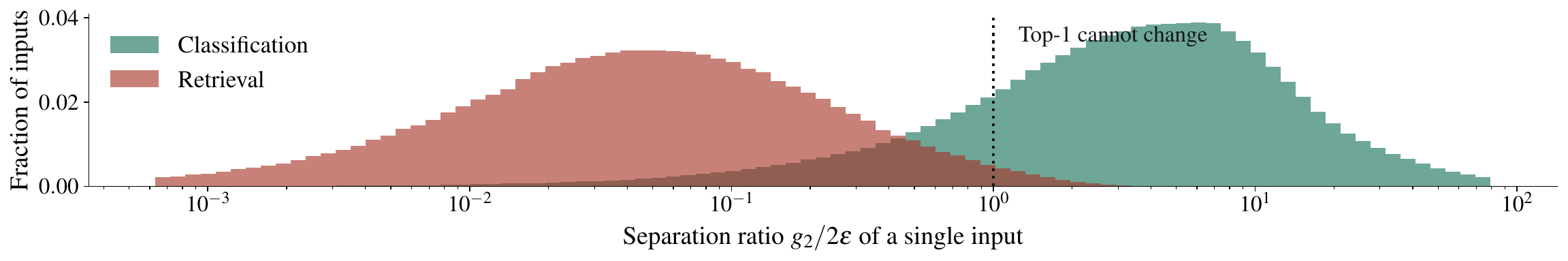}
\vspace{-0.7em}
\caption{\textbf{The two task families separate their winners by different amounts.} The separation ratio of every input we measure, at W4-group\_128, over both modalities. The dotted line is the stability threshold.}
\label{fig:sephist}
\end{figure}

The contender set $\cset$ is what separates the two tasks. It consists of the candidates within $2\varepsilon$ of the top-1, and at W4 it is empty for $74.3$\% of ViT-B/16 classification inputs but for none of the retrieval queries. The flips themselves are also informative. When classification changes its top-1, the new top-1 is the old runner-up $82.2$\% of the time, while in image retrieval it is only $39.7$\%, with the rest coming from deeper in the ranking, as a wider contender set predicts. The same holds on \textsc{clip} (Table~\ref{tab:clip-mechanism}). Over every input, $85.8$\% of classification inputs sit above the stability threshold against only $3.5$\% of retrieval queries, with median separation ratios of $5.73$ and $0.081$ (Figure~\ref{fig:sephist}). Among inputs with $|\cset| > 1$, the change rate rises with $|\cset|$ in both tasks, see Table~\ref{tab:pflip}.

\begin{wraptable}[17]{r}{0.5\linewidth}
\vspace{-1.2em}
\centering
\scriptsize
\caption{\textbf{The stability check on classification, text retrieval and \textsc{clip}.} Coverage (mean $\pm$ s.d. over $20$ calibration splits) and conditional violation rate at W4 group\_128. The thresholds themselves are in Table~\ref{tab:certificate-tau}.}
\label{tab:certificate}
\vspace{-0.5em}
\input{tables/new_certificate}
\end{wraptable}

\begin{figure}[t]
\centering
\vspace{-0.5em}
\includegraphics[width=0.9\linewidth]{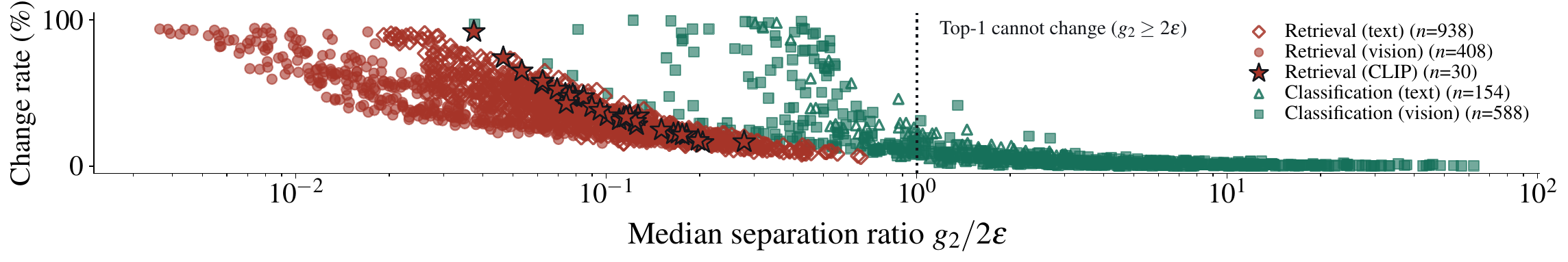}
\vspace{-0.7em}
\caption{\textbf{One quantity predicts the top-1 change rate in both task families.} Each point is one setup (model, task, quantizer, bit-width), and a retrieval setup appears once per corpus size.}
\label{fig:law}
\vspace{-1em}
\end{figure}

The median separation $\gapk{2}/2\varepsilon$ tracks the change rate at Spearman $-0.88$ over $2{,}118$ configurations, every one we measure (Figure~\ref{fig:law}). Within each family, the correlation is $-0.86$ (classification, text), $-0.92$ (classification, vision), $-0.93$ (retrieval, text), and $-0.90$ (retrieval, vision). Candidate count reaches only $+0.02$ on text and $+0.28$ on vision. The relationship survives intervention, which correlation alone cannot show. We fit a straight line of change rate against $\log_{10}$ of the median separation ratio by least squares. Fitted on the \rtn{} image-retrieval points alone at W4 ($n=34$, $R=-0.860$), the line predicts the same $34$ points under each of the five other quantizers, which it never saw, at $R^2$ between $0.60$ and $0.74$. Changing the quantizer moves both quantities but keeps their relationship. The ratio divides by $\varepsilon$, and $\varepsilon$ is also what causes the flips, so the correlation could seem trivial. It is not: on a single input, a changed top-1 always has $\gapk{2} < 2\varepsilon$, so Figure~\ref{fig:law} measures how well the gap predicts the damage, and neither quantity alone is as predictive. Alone, the median gap $\gapk{2}$ reaches Spearman $-0.58$ and the median perturbation $\varepsilon$ just $+0.05$, against $-0.88$ for the ratio ($\gapk{2}/2\varepsilon$).

The stability check of \S\ref{sec:theory}, which fixes the threshold in advance, is the deployable version that can fail, and Table~\ref{tab:certificate} reports its violations. On text retrieval, it accepts about one query in nine, since the separation there is two orders of magnitude smaller, and on \textsc{clip} the outcome is similar (Table~\ref{tab:certificate}). The violation rate stays within $\alpha$ in every calibration split. This was not guaranteed because $\hat\tau$ is fixed on calibration data, so the rate could have exceeded $\alpha$ on test data (Appendix~\ref{app:certificate}).

% =============================================================================
\subsection{The same effect in other systems}\label{sec:systems}

\textit{CLIP} text-to-image retrieval, a fourth architecture family~\citep{radford2021clip,young2014flickr30k}, shows the failure on a deployed system: captions as
queries, images as the corpus. ViT-L/14 at W4 loses only $1.3$\% of its
Recall@1, so the benchmark says quantization did little damage, while $24.6$\% of top-1 results
change and one right answer in nine becomes wrong (Table~\ref{tab:clip}). The metric stays flat because almost as many queries gain their correct
image as lose it. The queries that keep it are the ones with the larger separation, and they
change their top-1 four times less often.

\textit{Reranking}, the second stage of a search system, breaks the same way. A cross-encoder takes each
query and document together and rescores the shortlist the \fp{} retriever returned, and we
quantize only that cross-encoder~\citep{nguyen2016msmarco} while keeping the shortlist fixed. The
reranked top-1 changes for $5$--$10$\% of queries at W4 and $12$--$27$\% at W3, and the settings
with more separation change fewer results (Table~\ref{tab:crossencoder}). %Nothing in the previous analysis covered this architecture, and the same failure still emerged.

A \textit{retrieval-augmented} pipeline returns a different answer to the user once its retriever is
quantized, and at W4 its accuracy reports a twentieth of that change.  We quantize only the query encoder, Qwen3-Emb-0.6B, and leave the generator, Qwen3-8B, at \fp{} with greedy decoding, so every changed answer comes from retrieval, and we score $2{,}255$ Natural Questions~\citep{kwiatkowski2019natural} over $2.68$M
passages (results in Table~\ref{tab:rag}). The retrieved passages carry the answer, since the same
generator scores $15$\% with no passages against $43$\% with five. At W4, between $24$ and $32$\% of the answers change under the six weight quantizers, while exact match by at most $1.6$ points. More answers turn wrong than right, so the damage
is real and the metric understates it. We report W4 only, because at W3 this encoder returns
near-random passages and the pipeline does worse than the generator alone without even retrieval augmentation, which
makes that setting unusable. 

\textit{Encoder-based embedders} follow the same rule, and separation orders models better than size does. With three encoder-based embedders~\citep{xiao2024cpack,wang2022e5,li2023gte} beside the three Qwen3 models, the separation ratio predicts the change rate at Spearman $-0.98$ over the twelve model and bit-width configurations (Table~\ref{tab:arch}), and GTE-large, with a twelfth of the parameters of Qwen3-Emb-4B, changes fewer top-1 results at both bit-widths. Thus, one should pick higher-separation models for retrieval, not larger ones.

% =============================================================================

\subsection{Different quantization methods do not remove the asymmetry}\label{sec:quantizers}

\begin{figure}[t]
\centering
\includegraphics[width=0.95\linewidth]{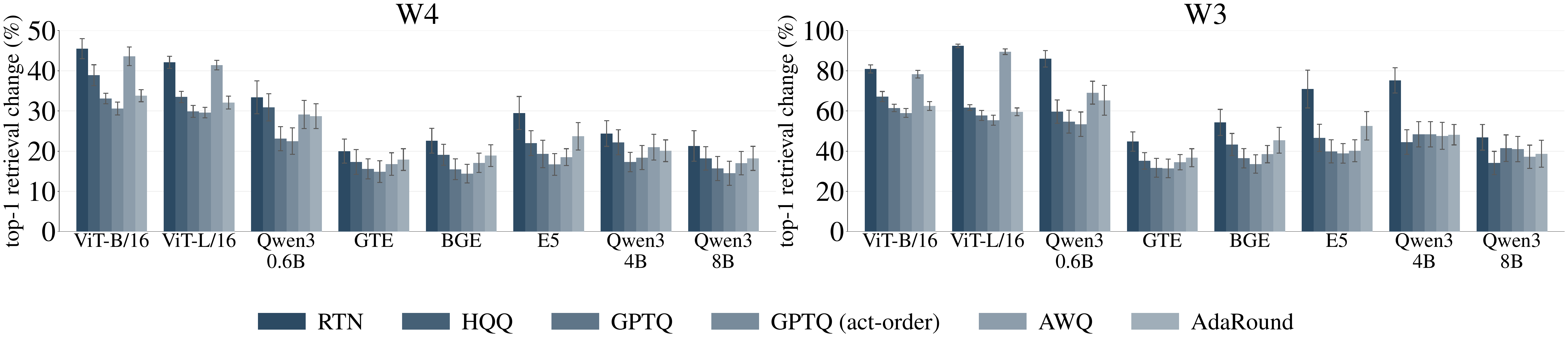}
\vspace{-1em}
\caption{\textbf{Stronger quantizers reduce ranking damage and do not remove it.} Top-1 retrieval change rate at group\_128, mean $\pm$ standard error over tasks. Numbers are in Table~\ref{tab:quantizers}.}
\label{fig:quantizers}
\vspace{-1.5em}
\end{figure}

The analysis so far is based on the \rtn{} quantizer, and an objection could be that it is too weak. We therefore repeat the measurement with five others (Figure~\ref{fig:quantizers}). \gptq{}~\citep{frantar2023gptq} compensates rounding error against the activation covariance, with and without activation ordering, \awq{}~\citep{lin2024awq} rescales to protect outlier channels, HQQ~\citep{badri2023hqq} fits scale and zero-point under a robust loss, and AdaRound~\citep{nagel2020adaround} learns the rounding direction of every weight. Settings are detailed in Appendix~\ref{app:setup-detail}, \emph{Quantizers}. All five reduce the top-1 change rate, including on \textsc{clip} (Table~\ref{tab:regime-clip}), but none of them approaches the stability threshold on retrieval. 

A finer quantization grid also does not solve the issue. We hold the bit-width fixed and change only how many weights share a scale, from one per output channel to one per $128$ weights to one per $64$. The rounding error falls, and the separation ratio $\gapk{2}/2\varepsilon$ rises under all six quantizers and in both task families, but the asymmetry widens (Table~\ref{tab:granularity}). Moving from the coarsest grid to the finest cuts the classification change rate by $3.7\times$, and the retrieval rate by $1.7\times$, so retrieval ends up changing $4.5$ to $8.0\times$ more of its top-1 results than classification, against $1.7$ to $5.9\times$ at the coarsest grid. A finer grid buys a real improvement in both families but moves neither across the threshold between them.

Quantizing the activations as well changes the size of the perturbation, but our results still hold. We add eight-bit activations, with one static range per tensor, on top of W4 and W8, under all six weight quantizers (Appendix~\ref{app:activation}). The change rate still rises
the same way, and retrieval's top-1 results still change seven to eight times
more than classification. At W8A8, still no retrieval configuration clears the stability threshold while $91$\% of classification configurations do. A configuration's typical input does not flip when its median separation $\gapk{2}/2\varepsilon \ge 1$. At W3, \gptq{} with and without activation ordering, HQQ and AdaRound push classification over the stability threshold, \rtn{} and \awq{} do not, and all six leave retrieval far below it. \gptq{} on ViT-L/16 restores accuracy from 55.5\% to 88.6\% and cuts the change rate eleven-fold, while retrieval on the same model still changes most of its top-1 results (Table~\ref{tab:regime}). Embedding retrieval sits one to two orders of magnitude below, too far for any quantizer to push it over.

% =============================================================================
\section{Two fixes from one quantity}\label{sec:interventions}

The gap suggests two fixes, and one number decides between them: the \emph{at-risk fraction}, the share of inputs whose top-1/top-2 gap is below $2\varepsilon$. At W4, it is $94$--$100\%$ of retrieval queries but only $9$--$26\%$ of classification inputs. Retrieval has nearly all inputs at risk, so in a mixed-precision setup, we allocate more bit-width to the layers whose quantization
moves the gap most. Classification has only a minority at risk, and that minority can be detected without labels. So the fix is per input, and we route those inputs to the \fp{} model and leave the rest on the quantized one. Both fixes need the same quantity, measured before deployment on unlabeled data.

\subsection{Retrieval: allocating bits by gap sensitivity}\label{sec:allocation}

Weight-only PTQ methods decide \emph{how} to round and not
\emph{where} to spend precision. Bit-width is set once globally for all layers, and the goal is to minimize reconstruction error. Where to spend it is a separate choice, and the two combine freely: an allocation criterion decides which layers get more bits, and any quantizer then rounds them, so a criterion plugs into every quantizer. We propose an allocation criterion and compare it against baselines, under each quantizer fixed. Fix the mixed-precision \emph{budget} at $3.5$ bits, the average number of bits per weight the deployment allows. The question is, which half of the layers receive four bits and which receive three?

\textbf{Method.} Run the \fp{} model on the $n = 128$ calibration queries against the corpus. For each query, this gives the top-1/top-2 gap $\gapk{2}(x)$ and the identities of the two documents involved. Then, for each linear layer $\ell$, quantize that layer alone, with plain round-to-nearest (RTN) at 3 bits and group size 128, leave every other layer at \fp{}, rerun the calibration queries, and compute the score difference between the \emph{same two documents}, $\gapk{2}^{(\ell)}(x)$. The sensitivity of the layer is how much this one-layer quantization moves the gap on a typical query,

\[
  s_\ell \;=\; \mathrm{median}_{x}\bigl|\,\gapk{2}(x) - \gapk{2}^{(\ell)}(x)\,\bigr|.
\]

Layers are ranked by $s_\ell$ divided by their parameter count, since bits are spent per weight, and the higher bit-width goes to the most sensitive layers until the budget is spent. The whole measurement is $L+1$ forward passes over $128$ queries for a model with $L$ linear layers, needs no labels, and trains nothing. The sensitivities are measured once, with RTN, and the resulting assignment is then applied with whatever quantizer the deployment uses, so the method generalizes to unseen quantizers.

\textbf{Baselines and metric.} We compare gap sensitivity against allocating by reconstruction error, the objective of existing allocators, and five other criteria, defined in Appendix~\ref{app:alloc-criteria}. We report the fraction of the $\mathrm{W}3 \to \mathrm{W}4$ benefit that an allocation captures while spending only half the extra bit at $3.5$ bits, computed as $\sum \text{gain} / \sum \text{headroom}$ (the \emph{capture ratio}).

\begin{figure}[t]
\centering
\includegraphics[width=0.95\linewidth]{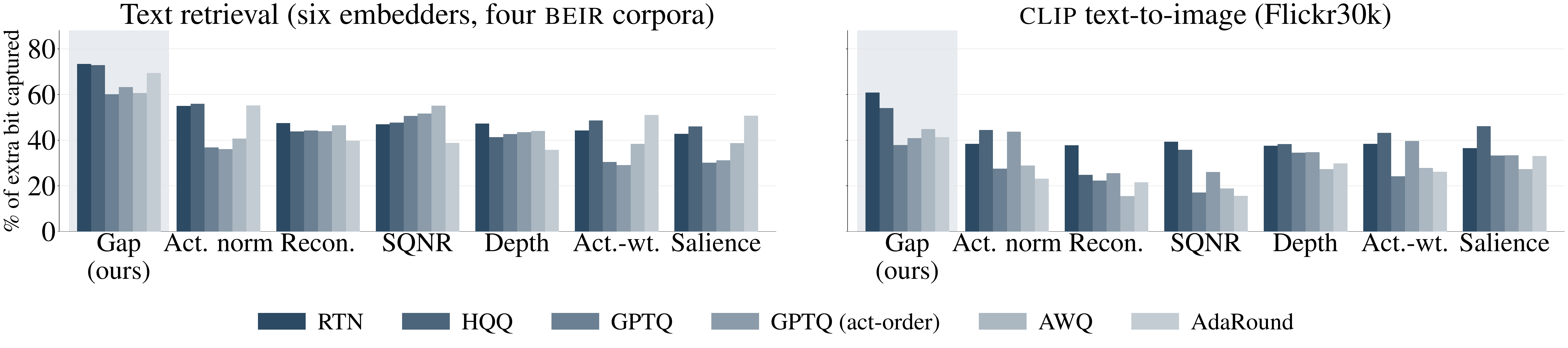}
\vspace{-1em}
\caption{\textbf{Gap sensitivity outperforms all other criteria we compare.} Fraction of the W3$\to$W4 reduction in top-1 change recovered at a $3.5$-bit average, averaged across models and corpora.}
\label{fig:criteria}
\vspace{-1em}
\end{figure}

\textbf{Gap sensitivity leads.} Gap allocation recovers $60$--$73$\% of the extra bit's benefit 
and is first among all seven forward-only criteria under every quantizer (Figure~\ref{fig:criteria}, Table~\ref{tab:alloc-nograd}). Its lead over the next best criterion ranges from $5.5$ points (\awq{}) to $18.4$ (\rtn{}), and it beats each competitor in at least $16$ of the $24$ configurations. In absolute terms, at the $3.5$-bit budget under \rtn{}, it changes $33.3$\% of top-1 results against $43.1$\% for reconstruction error and loses $13.8$\% of correct top-1 documents against $19.9$\% (Table~\ref{tab:alloc-absolute}), a mean paired gain of $+26.2$ capture points over the $24$ configurations. Reconstruction error, the objective of existing allocators, captures only $40$--$48$\%. The reason is visible at the layer level: on the text embedders, reconstruction error is nearly uncorrelated with gap sensitivity across layers (mean Spearman $-0.04$, Table~\ref{tab:alloc-permodel}), and on \textsc{clip}, where the correlation reaches $+0.34$, it recovers more but still less than gap sensitivity.

\textbf{Retrieval-augmented generation.} Gap allocation also changes fewer answers under every quantizer. We run the same retrieval-augmented pipeline on Natural Questions at a $3.5$-bit budget, change only the allocation, and repeat it for all six quantizers
(Table~\ref{tab:alloc-rag}). We compare gap allocation against reconstruction error, the most common objective that existing allocators optimize. On average, gap sensitivity recovers more of the extra bit under all quantizers by $18$pp and leaves $62$ fewer answers changed. Hence, gap allocation yields higher agreement between the quantized and \fp{} models.

\textbf{Robustness.} The lead holds under every setting we vary. Across the six models, per-model capture stays between $68.3$\% and $84.5$\% (Table~\ref{tab:alloc-permodel}). It is unchanged at the other two budgets we test, $3.25$ and $3.75$ bits per weight, across calibration seeds, and at every calibration size from $8$ to $256$ queries, on text and on \textsc{clip} (Appendix~\ref{app:alloc-robust}). On \textsc{clip} itself, over both encoders and all seven settings of budget, seed and quantizer, gap-driven allocation captures $56.4$\% of the extra bit against $34.9$\% for reconstruction error, in every configuration (Table~\ref{tab:alloc-clip}, full results in Appendix~\ref{app:clip}). The lead also survives quantizing the activations: with eight-bit activations on top of weight quantization, gap sensitivity captures $60$ to $76$\% of the extra bit against $33$ to $53$\% for reconstruction error under each of the six quantizers (Appendix~\ref{app:activation}).

\subsection{Classification: routing inputs by the gap}\label{sec:routing}

In classification, the same gap that gives the stability check also provides a fix. Routing low-gap inputs to the \fp{} model recovers most of the lost accuracy at a fraction of the \fp{} compute.

\textbf{Method.} Let $\qmargin(x)$ be the top-1/top-2 gap of the \emph{quantized} logits. It comes at no extra cost, from the \ptq{} forward pass the deployment already runs. By the top-1 condition of Lemma~\ref{lem:cuts}, $\qmargin(x) > 2\varepsilon$ means no
perturbation of size $\varepsilon$ changes the prediction, so at every budget the flippable
inputs are exactly those with $\qmargin(x) < 2\varepsilon$: one threshold on the gap
separates them. Route $x$ to \fp{} iff $\qmargin(x) < \tau$, with
$\tau$ set at the 25th percentile of \qmargin{} on an unlabeled validation slice: no
training, just one tuned hyperparameter. Up to order-equivalence, the gap is the only score of the quantized logits whose threshold sets match the worst-case flippable sets at every budget (Proposition~\ref{prop:routing}).

\textbf{Results.} Routing $25\%$ of inputs recovers $85$--$93\%$ of the accuracy lost to quantization on the three fine-tuned backbones, over the tasks whose \fp$\to$\ptq{} accuracy gap is at least $0.5$ points (Table~\ref{tab:routing-sweep}). The cost is one quantized pass over everything plus one \fp{} pass over the routed quarter, $47\%$ of the \fp{} cost at the $4.5\times$ speedup \gptq{} reports~\citep{frantar2023gptq}. Recovery rises smoothly with the fraction routed, from $62$--$76\%$ at $10\%$ to $97$--$99\%$ at $50\%$ (Table~\ref{tab:routing-sweep}). At nearly every operating point, the gap recovers more than \textsc{msp} or predictive entropy~\citep{gal2016dropout}.

\section{Conclusion}

We show that the damage caused by quantization is tied to the score gaps between the top two candidates (logits in classification, query-document similarities in retrieval). When that gap is large, as in classification, the quantized model preserves its accuracy on most inputs. When the gap is small, as in retrieval, the quantized model changes many top-1 results, and standard metrics understate that damage. 
The gap is computable label-free and predicts which models will break before quantization and which answers to trust after it. The gap also motivates a fix in each task. In retrieval, spending extra bit-width on the layers whose quantization moves the gap most recovers most of an extra bit's benefit for half its cost. In classification, routing the few low-gap inputs to \fp{} recovers most of the lost accuracy at a fraction of the \fp{} cost. We advocate that compression should be judged beyond standard metrics, and the open question is whether training can encourage the gap that a target task benefits from, making a model more robust to quantization under specific deployment needs.

\section*{Acknowledgements}
This work is supported by the MUR FIS2 grant n. FIS-2023-00942 "NEXUS" (cup B53C25001030001), and partly by Sapienza University of Rome via the Seed of ERC grant "MINT.AI" (cup B83C25001040001).

\section*{Reproducibility statement}\label{sec:repro}

Every number in this paper comes from code we release, and every table
and figure is generated from stored result files. The theory is
self-contained. Lemma~\ref{lem:cuts} and the statements it rests on are proved in
Appendix~\ref{app:proofs} and checked by exhaustive enumeration (Appendix~\ref{app:theory-detail}, \emph{Boundary cases and verification}). Section~\ref{sec:setup-sub} gives
the models, datasets, quantization settings and calibration protocol. All calibration is label-free, and the calibration protocol in Appendix~\ref{app:setup-detail} states, per procedure, which calibration splits are disjoint from evaluation and where retrieval overlaps. The
supplementary material contains the code and the stored result files needed to regenerate every
table and figure in the paper.

\section*{Ethics statement}

This work studies the reliability of an existing deployment practice and introduces no new
models, data collection, or human subjects. All datasets are public, standard benchmarks used
under their original licenses. The practical consequence we report is a safety one. Quantized retrieval systems can degrade in ways their accuracy metrics do not reveal, which may disadvantage users whose queries fall in the affected minority. The stability check and the allocation criterion are offered to make that risk measurable and reducible before deployment.
We report the settings in which our stability check fails alongside those in which it holds.

\section*{Use of large language models}

Large language models were used as a general-purpose coding and writing assistant: drafting and
refactoring experiment and plotting scripts, and editing prose for concision. All research
questions, experimental designs, analyses, and claims are the authors' own, and the authors
verified every reported number against the stored result files.

\section*{Limitations}
Our results cover weight quantization at 4 and 3 bits (W4, W3), and eight-bit activations on top of W4 and W8 with one static range per tensor.
Finer activation schemes such as per-token ranges are untested. The downstream evidence of RAG damage is one pipeline: a Qwen3-Emb-0.6B retriever with a Qwen3-8B generator on Natural Questions at W4. The stability check assumes that calibration and deployment inputs come from the same distribution, the standard conformal assumption, and will degrade under distribution shift. Retrieval results use exact cosine similarity over a finite corpus. Approximate nearest-neighbor indexes, which add ranking errors of their own, are out of scope.

\bibliographystyle{iclr2027_conference}
\bibliography{references}

% =============================================================================
\appendix
\section{Additional results and experimental detail}\label{app:additional}

This appendix collects the material referenced from the main text: the tables cited by number (\S\ref{app:tables}), the datasets, the retrieval construction and the full calibration protocol (\S\ref{app:setup-detail}), ranking damage beyond the top-1 (\S\ref{app:beyond-top1}), corpus size and model scale (\S\ref{app:corpus-scale}), the stability check against confidence baselines and the predicted flip curve (\S\ref{app:certificate}), the \textsc{clip} arm in full (\S\ref{app:clip}), the allocation criteria we compare against (\S\ref{app:alloc-criteria}), the robustness of the allocation to budgets and calibration seeds (\S\ref{app:alloc-robust}), and activation quantization (\S\ref{app:activation}).

\subsection{Tables referenced from the main text}\label{app:tables}
\begin{table}[t]
\centering
\footnotesize
\caption{\textbf{A deployed cross-modal retrieval system.} \textsc{clip} text-to-image on Flickr30k, $4000$ images, $2000$ caption queries, zero-shot. The relevant image for a caption is the one it was written for. Recall@1 is the metric the system is judged by. Parenthesized values are relative changes in Recall@1. $^{\ast}$Fraction of the queries whose correct image was ranked first under \fp{} and no longer is.}
\label{tab:clip}
\input{tables/new_clip}
\end{table}
A note on Table~\ref{tab:clip}. At W3, ViT-L/14 loses a third of its Recall@1, and ViT-B/32 collapses from $50.1$\% to $9.4$\%. The smaller encoder is more fragile at both bit-widths.
\begin{table}[t]
\centering
\footnotesize
\caption{\textbf{Among allocation criteria that need only forward passes, gap sensitivity is first under every quantizer.} Fraction of the W3$\to$W4 reduction in top-1 change recovered at a $3.5$-bit average, pooled over models and corpora. Every criterion here is computable inside a serving deployment: no backward pass, no autograd, no calibration loss to choose.}
\label{tab:alloc-nograd}
\input{tables/new_alloc_nograd}
\end{table}
\begin{table}[t]
\centering
\small
\caption{\textbf{The same comparison on the deployed cross-modal system.} \textsc{clip} text-to-image retrieval on Flickr30k (ViT-B/32 and ViT-L/14, zero-shot): fraction of the W3$\to$W4 reduction in top-1 change recovered at a $3.5$-bit average, pooled over both encoders, one calibration seed per quantizer. Calibration captions are held out of evaluation. Gap sensitivity leads every forward-only criterion under every rounding method. It wins 6 of 6 runs against reconstruction error.}
\label{tab:alloc-clip}
\input{tables/new_alloc_clip}
\end{table}

\begin{table}[t]
\centering
\footnotesize
\caption{\textbf{The asymmetry survives the strictest control.} One encoder (Qwen3-Emb-0.6B), one dataset, one quantization: the \emph{same} embeddings of the test split are used twice, as classification through a linear probe fitted on full-precision training embeddings and then frozen, and as retrieval, by cosine nearest neighbor within the split. Only the task differs, so the gap between the columns cannot be attributed to a different checkpoint, dataset, or objective. Separation is the median $\gapk{2}/2\varepsilon$ of the corresponding scores.}
\label{tab:contrast}
\input{tables/new_contrast}
\end{table}
\begin{table}[t]
\centering
\footnotesize
\caption{\textbf{The separation ratio orders models across architectures, not only within one.} Top-1 retrieval change and median separation over the four text corpora. Six models, two architecture families, a $24\times$ parameter range. Across all 12 model--bit-width configurations the separation ratio predicts the change rate at Spearman $-0.979$ ($p=3e-08$). Restricted to W3, $-0.943$ over six models.}
\label{tab:arch}
\input{tables/new_arch}
\end{table}
\begin{table}[t]
\centering
\footnotesize
\caption{\textbf{A quantized cross-encoder rewrites the reranked order, and the separation ratio
predicts by how much.} A full-precision bi-encoder retrieves a
fixed top-100 shortlist per query. Only the cross-encoder reranker
(\texttt{ms-marco-MiniLM-L-6-v2}) is quantized. 500 queries per corpus. Flip is the fraction of
queries whose reranked top-1 changes. Across the six configurations the separation ratio orders the flip
rate at Spearman $-0.943$ ($p=0.005$).}
\label{tab:crossencoder}
\input{tables/new_crossencoder}
\end{table}

\begin{table}[t]
\centering
\footnotesize
\setlength{\tabcolsep}{3.5pt}
\caption{\textbf{Stronger quantizers reduce ranking damage and do not remove it.} Top-1 retrieval change rate (\%) at group\_128, mean $\pm$ standard error over tasks.}
\label{tab:quantizers}
\input{tables/new_quantizers}
\end{table}
\begin{table}[t]
    \centering
    \scriptsize
    \caption{\textbf{Separation, not change rate alone, shows which configurations cross the stability threshold.} Median separation ratio $\gapk{2}/2\varepsilon$, top-1 change rate and quantized accuracy at group\_128, both ViTs, six quantizers.}
    \label{tab:regime}
\input{tables/new_regime}
\end{table}

\begin{table}[t]
\centering
\footnotesize
\caption{\textbf{Absolute retrieval metrics under each allocation.} Mean over the 24-configuration headline grid (six models $\times$ four corpora, RTN, average 3.5 bits for the middle rows). Flip is disagreement with the full-precision top-1. Recall@1 and gold retention are against relevance judgements. The capture ratios of Table~\ref{tab:alloc-nograd} are computed from these absolute values.}
\label{tab:alloc-absolute}
\input{tables/new_alloc_absolute}
\end{table}

\begin{table}[t]
\centering
\footnotesize
\setlength{\tabcolsep}{4pt}
\begin{minipage}[t]{0.49\linewidth}\centering
Gap sensitivity (ours)\\[2pt]
\input{tables/new_alloc_rag_gap}
\end{minipage}\hfill
\begin{minipage}[t]{0.49\linewidth}\centering
Reconstruction error\\[2pt]
\input{tables/new_alloc_rag_mse}
\end{minipage}
\caption{\textbf{Gap allocation keeps the pipeline closer to \fp{} under every quantizer.} Natural Questions at a $3.5$-bit average, against reconstruction error, the objective existing allocators optimize. The other criteria are compared in Figure~\ref{fig:criteria}. \emph{Capture} is the share of the W3$\to$W4 reduction in top-1 change recovered, over $3{,}324$ queries on a $100{,}000$-passage subsample. \emph{Changed} counts, of the $2{,}255$ questions that have a gold short answer, how many the pipeline answers differently from \fp{}, and \emph{R$\to$W} how many of those turn from correct to incorrect, with a \fp{} Qwen3-8B generator reading five passages. Gap sensitivity is better on every measure under every quantizer.}
\label{tab:alloc-rag}
\end{table}

\begin{table}[t]
\centering
\footnotesize

\caption{\textbf{Per model: gap-driven allocation wins on every one, and reconstruction error does not track gap damage.} Capture of the W3$\to$W4 flip-rate benefit at a 3.5-bit average under \rtn{}, per model, pooled over the four corpora, and the Spearman correlation between per-layer reconstruction error and per-layer gap sensitivity, mean over all \rtn{} allocation runs with the range across runs and the number of layers $n$.}
\label{tab:alloc-permodel}
\input{tables/new_alloc_permodel}
\end{table}
\begin{table}[t]
    \centering
    \scriptsize
    \caption{\textbf{Routing score and threshold sensitivity.} Fraction of inputs routed and fraction of the quantization accuracy loss recovered, at three percentiles of the routing score on an unlabeled validation slice, W4-channel, averaged over the eligible tasks per backbone, those whose \fp$\to$\ptq{} accuracy gap is at least $0.5$ points (count in parentheses).}
    \label{tab:routing-sweep}
\input{tables/new_routing_sweep}
\end{table}

\FloatBarrier

\begin{table}[t]
\centering
\footnotesize
\setlength{\tabcolsep}{4pt}
\caption{\textbf{A finer quantization grid helps both task families and separates them further.} Top-1 change rate at each scale granularity, W4 and W3 pooled, paired over the $2{,}004$ configurations measured at all three.}
\label{tab:granularity}
\input{tables/new_granularity}
\end{table}

\begin{table}[t]
\centering
\scriptsize
\setlength{\tabcolsep}{4pt}
\caption{\textbf{Quantizing the retriever changes answers that the benchmark does not report.} Retrieval-augmented generation on Natural Questions, $2{,}255$ questions over $2.68$M passages,  with the index embedded at \fp{}, only the query encoder Qwen3-Emb-0.6B quantized, and a \fp{} Qwen3-8B generator decoding greedily over five passages, one row per weight quantizer at W4}. The last three columns count questions.
\label{tab:rag}
\input{tables/new_rag}
\end{table}

\subsection{Datasets, retrieval construction and calibration}\label{app:setup-detail}
\paragraph{Models and settings.} The ViTs use the \texttt{orig\_in21k} initialization via \texttt{timm}~\citep{rw2019timm}, and the fine-tuning protocol follows the task-arithmetic literature~\citep{ilharco2023editing,ilharco2022patching}. The classification head is excluded from quantization throughout, and the five quantizers other than \rtn{} are evaluated at the group granularity. Every ranking is by cosine similarity between L2-normalized embeddings. The ViTs use the pooled representation just before the classification head. The text embedders use their own pooling through \texttt{sentence-transformers}~\citep{reimers2019sbert}, which for Qwen3-Embedding is last-token pooling with the instruction prefix these models expect on queries. \textsc{clip} uses its image and text projections. \textsc{clip} is the OpenAI ViT-B/32 and ViT-L/14 release, loaded through \texttt{open\_clip}: an image ViT (12 blocks of width 768 on 32-pixel patches, or 24 blocks of width 1024 on 14-pixel patches) and a 12-block text Transformer (width 512 or 768), each followed by a linear projection into a shared space of 512 or 768 dimensions. Ranking is by cosine similarity, and there is no fine-tuning. Every linear layer of both encoders is quantized.

\paragraph{Quantizers.} All six are weight-only and quantize every linear layer. \rtn{} rounds to the nearest grid point with a symmetric absmax scale per channel or per group. \gptq{} uses block size $128$ and $1$\% dampening of the Hessian diagonal, and its activation-ordered variant quantizes columns in decreasing order of that diagonal. \awq{} searches the scaling exponent on a grid of $20$ values. HQQ fits scale and zero-point under an $\ell_p$ loss with $p = 0.7$ by half-quadratic splitting, $20$ iterations from a min-max initialization. AdaRound keeps the \rtn{} scale and learns each weight's rounding direction with the rectified sigmoid of the original paper, $400$ Adam steps at learning rate $0.03$, regularizer weight $10$, annealed from $\beta = 20$ to $2$ after a $20$\% warm-up. Its reconstruction loss is computed exactly from the calibration second moment $X^\top X$ rather than from minibatches. \gptq{}, \awq{} and AdaRound calibrate on $512$ documents and quantize layers sequentially, each seeing the outputs of the already quantized layers before it. HQQ and \rtn{} are data-free.

\paragraph{Datasets.} The 21 image classification tasks are
Cars~\citep{krause_3d_2013}, CIFAR-10 and CIFAR-100~\citep{krizhevsky_learning_nodate},
DTD~\citep{cimpoi_describing_2014}, EMNIST~\citep{cohen_emnist_2017},
EuroSAT~\citep{helber_eurosat_2019}, FashionMNIST~\citep{xiao_fashion-mnist_2017},
FER2013~\citep{goodfellow_challenges_2013}, Flowers102~\citep{nilsback_automated_2008},
Food101~\citep{bossard_food-101_2014}, GTSRB~\citep{stallkamp_german_2011},
KMNIST~\citep{clanuwat_deep_2018}, MNIST~\citep{noauthor_mnist_nodate},
OxfordIIITPet~\citep{parkhi_cats_2012}, PCAM~\citep{veeling_rotation_2018},
RenderedSST2~\citep{socher_recursive_nodate}, RESISC45~\citep{cheng_remote_2017},
STL10~\citep{coates_analysis_2011}, SUN397~\citep{xiao_sun_2016},
SVHN~\citep{netzer_reading_nodate} and TinyImageNet~\citep{le2015tiny}. The 11 text
classification tasks come from \textsc{mteb}~\citep{muennighoff2023mteb}:
AmazonCounterfactual~\citep{oneill-etal-2021-wish},
AmazonReviews~\citep{keung2020multilingual}, Banking77~\citep{casanueva-etal-2020-efficient},
Emotion~\citep{saravia-etal-2018-carer}, IMDB~\citep{maas-etal-2011-learning},
MassiveIntent, MassiveScenario~\citep{fitzgerald2022massive},
MTOPDomain, MTOPIntent~\citep{li-etal-2021-mtop},
ToxicConversations~\citep{jigsaw-unintended-bias-in-toxicity-classification} and
TweetSentimentExtraction~\citep{tweet-sentiment-extraction}.

\paragraph{Image self-retrieval.}
The image setting is self-retrieval: the corpus
is the task's test split, queries are drawn from it, and the self-match is removed before ranking.
It is a controlled setup, not a deployed system, and we use it on purpose. Because the encoder
is the \emph{same} fine-tuned checkpoint that produces the classification results, it isolates the one thing that changes: the classification and retrieval numbers in
Table~\ref{tab:fragility} differ only in how the output is read, not in the model, the data, the
quantizer, or the training. No standard retrieval benchmark offers that control. The text setting provides what this setup cannot (real corpora, real queries, and relevance judgements), and the two are always reported separately, never pooled.

\paragraph{Calibration protocol.} Six procedures take calibration data (the allocation criteria, \gptq{}, \awq{}, AdaRound, routing, and the stability check) and none of them takes
labels. The allocation criteria are estimated from at most $128$ queries, capped at a quarter
of the query pool, or in classification from $4$ batches. \gptq{}, \awq{} and AdaRound collect activation statistics from $512$
documents. Routing sets its threshold on an unlabeled validation slice. The stability check
calibrates its threshold on a conformal split. Calibration and evaluation are disjoint throughout.
In classification, the calibration inputs come from the \emph{train} split while evaluation is on
test. In retrieval, the calibration queries are a seeded random slice held out of the query pool,
and every rate we report is computed on the complement. Routing and the stability check likewise fix
their thresholds on held-out slices and report on the remainder. One overlap remains, on the
document side: \gptq{}, \awq{} and AdaRound draw their calibration documents from the indexed corpus, as is
standard for post-training quantization. Nothing that is evaluated is selected using them, since
the quantizer never sees a query.

One pair of embeddings serves the whole corpus-size sweep, since restricting the corpus only changes which scores are compared. The same pair also gives the query-only condition, in which the corpus is embedded offline in \fp{} and only the query encoder is quantized. For each input, we record the top-$K$ scores, with $K = 10$ for classification and $200$ for retrieval. Retrieval needs the longer list because a corpus places many candidates within $2\varepsilon$ of the winner.

\subsection{Ranking damage beyond the top-1}\label{app:beyond-top1}
\begin{table}[t]
\centering
\footnotesize
\caption{\textbf{The protection is specific to the argmax.} Median separation at the $k$-cut relative to the perturbation, $\big(\gapk{k+1}-\gapk{k}\big)/2\varepsilon$, and the fraction of inputs whose exact top-$k$ \emph{set} is unchanged. W4-group\_128.}
\label{tab:topk}
\input{tables/new_topk}
\end{table}
\begin{table}[t]
\centering
\footnotesize
\caption{\textbf{Accuracy damage does not predict ranking damage.} Spearman correlation across tasks between the accuracy a configuration loses and the fraction of inputs whose exact top-$5$ set it changes, computed \emph{within} each configuration so bit-width and granularity are held fixed. Pooled across configurations the correlation is $+0.834$, but that compares settings poor at both against settings good at both. Removing the configuration effect leaves $+0.327$, and within the deployed W4-group\_128 setting nothing is detectable. $^{*}$significant at $0.05$.}
\label{tab:accrank}
\input{tables/new_accrank}
\end{table}
\paragraph{The protection is specific to the top-1.} On classification, where accuracy looks nearly unchanged, the damage below the first position is already visible. The top-$k$ part of Lemma~\ref{lem:cuts} guarantees the top-$k$ set
when the gap between rank-$k$ and rank-$k+1$ items exceeds $2\varepsilon$. Table~\ref{tab:topk}
gives that gap divided by $2\varepsilon$ at every rank. Between
first and second place it is $2.4$ to $5.4$, so the top-1 cannot change. At every rank below, it falls to $0.5$ or less. Cross-entropy separates the
correct class from the rest and separates nothing else, so there is exactly one boundary the
stability check can use, which is why the one we deploy (\S\ref{par:certificate}) covers the top-1
only. On models whose top-1 accuracy is preserved, the exact
top-$5$ set survives for only $28$ to $47$\% of inputs. Preserved accuracy is only about the top two items and tells nothing about the
rest of the ranking.

\paragraph{Accuracy against ranking, in detail.} Table~\ref{tab:accrank} correlates, across tasks, the accuracy a configuration loses against the fraction of inputs whose top-$5$ set it changes. Within each quantization setting, the correlation is $+0.327$, about a tenth of the variance, and significant in only three of eight settings. In the deployed W4-group\_128 setting, it is undetectable on either vision backbone ($+0.089$, $p=0.72$ and $+0.230$, $p=0.34$). Accuracy preservation, as reported, is not evidence of ranking preservation.
\paragraph{Retrieval quality against relevance judgements.} Everywhere above, we measure agreement between the \fp{} and quantized rankings, on purpose: it needs no relevance labels, and it answers the deployment question directly. It leaves open whether the changed results are \emph{worse}. The text corpora come with relevance judgements, so we can answer that.

Its change rates are computed
over the judged queries only, those carrying relevance labels, and so differ slightly from the
all-query rates reported elsewhere. At W4 a fifth to a third of
top-1 results change for a $3.1$--$7.5$\% relative loss in nDCG@10, so the changes cost quality. The mean also hides the per-query outcome: of the flips whose \fp{} top-1 was a judged-relevant document, $67$\% replace it with one that is not (Table~\ref{tab:harm}). We repeat the distinction of \S\ref{sec:fragility} here because these numbers could be read the other way.

At W3, nDCG@10 falls by $19$--$69$\%, from a working retrieval system to one that returns mostly unrelated documents. At this bit-width the ranking is no longer usable. That is the practical meaning of two earlier findings: the stability check accepts nothing at W3, and no quantizer we tested repairs it.

A mean is the wrong statistic for this failure, and reporting one repeats the mistake this paper is about. The per-query question is instead: for how many
queries does a document that was both relevant and inside the \fp{} top-$k$ fall out of the
quantized top-$k$?

At the deployed setting, $8$B at W4, where mean nDCG@10 falls by only $3.1$\%, \textbf{$15.6$\% of the queries whose top result was relevant lose it}. One in six of those queries is harmed while the average moves by only three percent. At
$0.6$B the figures are $7.5$\% and $22.2$\%, and the lost fraction is within a point of these at $k = 5$ and $k = 10$. Averaging over queries understates per-query ranking damage, just as averaging over inputs understates it in accuracy.

Scale helps, and much more at W3 than at W4. Across a $13\times$ range of model size the W4 loss falls from $7.5$\% to $3.1$\%, with diminishing returns. The W3 loss falls from $68.9$\% to $18.9$\%: at $0.6$B three-bit retrieval is unusable, at $8$B it is degraded but usable. Quality improves faster than the change rate falls ($85.9$\% to $48.6$\% at W3), so larger models change fewer results, and a larger share of the changes they make are between documents of similar relevance.

\subsection{Does a flip replace a relevant document?}\label{app:harm}

\begin{table}[t]
\centering
\footnotesize
\caption{\textbf{A flip usually costs the relevant document, and the separation ratio predicts which
queries are affected.} Top-1 flips joined against the \textsc{beir} relevance judgements, over six
text embedders, four corpora and six quantizers, both encoders quantized on the full corpus
(73 runs at W4, 72 at W3). \emph{Harm} is the share of all queries in the bucket whose relevant
top-1 document is replaced by one that is not judged relevant. The text below reads the
table.}
\label{tab:harm}
\input{tables/new_harm}
\end{table}

Table~\ref{tab:harm} joins the top-1 flips of Table~\ref{tab:fragility} against the \textsc{beir}
relevance judgements. The corpus permutation and the query subsample are seeded, so the stored
document indices identify the retrieved documents and the join needs no re-encoding. A query enters
the table when it carries at least one judgement.

Pooled over the $6{,}578$ W4 flips, $14.6$\% replace a relevant document by one that is not,
$7.1$\% exchange one relevant document for another, $11.5$\% promote a relevant document that was
not first before, and the remaining $66.8$\% move between documents that carry no judgement at all.
The mean relevance of the served document therefore falls by $0.010$ at W4 and by $0.092$ at W3:
losses and gains partly cancel, the same pattern as the \textsc{clip} result in
\S\ref{sec:fragility}, where $6.5$\% of queries turn from right to wrong while $5.7$\% turn the
other way.

\subsection{Corpus size and model scale}\label{app:corpus-scale}
\begin{table}[t]
\centering
\footnotesize
\caption{\textbf{Corpus size matters only when it tightens the separation.} Top-1 change rate at the smallest and largest corpus, W4-group\_128, paired over queries ($p$ from McNemar's test). The final columns show why: adding images from the same label set drives the separation ratio $\gapk{2}/2\varepsilon$ down, while adding documents to an already diverse corpus does not.}
\label{tab:corpus}
\input{tables/new_corpus}
\end{table}
\paragraph{Corpus size, case by case.} Every comparison in Table~\ref{tab:corpus} is paired over the same queries. On the vision datasets, the separation ratio collapses as the corpus grows (\texttt{SUN397} $0.145 \to 0.041$), on \texttt{FiQA} it is flat, and on \texttt{SciFact}, the one dataset whose change rate falls, it rises ($0.091 \to 0.132$). On \textsc{clip}, ViT-B/32 is flat, and ViT-L/14 falls as the index grows, because the correct image enters it (Table~\ref{tab:corpus-clip}).
\paragraph{The deployable configuration is not safe.} A deployment stack embeds its corpus offline
in \fp{} and quantizes only the query encoder. Less of the system is quantized, so one might expect less damage. On the text corpora it helps modestly, reducing changes by
$3$--$10$ points, and on \textsc{clip} it helps most, from $47.0$\% to $27.1$\% on ViT-B/32 and from $24.6$\% to $15.3$\% on ViT-L/14 (Appendix~\ref{app:clip}, \emph{Corpus size}). On the vision backbones, it is often \emph{worse}. On ViT-L/16 it is worse on all five datasets, by up to $6.5$ points. Across every dataset, the top-1 result still changes
for $18.9$--$53.6$\% of queries. Keeping the index exact does not stabilize the ranking, because the perturbation that reorders the candidates acts on the query that ranks them, which an exact index does nothing to cancel.

\paragraph{Scale reduces the effect and does not remove it.} One might expect the problem to affect only small models. Across a $13\times$ range of parameters in one model family (Tables~\ref{tab:quantizers} and~\ref{tab:arch}), the top-1 retrieval change rate falls from $33.4$\% to $21.3$\% and the separation ratio roughly doubles. That is a real improvement, and it is in the direction the mechanism predicts: a larger model spreads its embeddings further apart relative to the same quantization noise.
The largest model still changes one query in five, three to six times the classification rate of the $0.6$B model (the only size in this family with fine-tuned classifiers), and its separation is still far below the threshold at which anything can pass. Scale, like a better quantizer, reduces the damage without removing it.

\subsection{Calibrating the stability check, and the predicted flip curve}\label{app:certificate}
\begin{table}[t]
\centering
\footnotesize
\caption{\textbf{The stability check against confidence scores under one protocol.} ViT-B/16, W4 group\_128, $21$ tasks, mean over $20$ calibration splits. Split conformal makes \emph{any} score valid, so every row respects its bound. The generic construction thresholds the score at its $(1-\alpha)$ quantile among flipped calibration inputs, which bounds the joint event for any score. What separates the scores is how many inputs each accepts at a given level.}
\label{tab:baselines}
\input{tables/new_baselines}
\end{table}

\begin{table}[t]
\centering
\footnotesize
\caption{\textbf{The stored thresholds of the stability check.} The value $\hat\tau$ that a test input's quantized top-1/top-2 gap must reach to be accepted, at each level $\alpha$, W4 group\_128, median over $20$ random calibration splits. Logits for classification, cosine similarity for the retrieval rows. Coverage and violation rates are in Table~\ref{tab:certificate}.}
\label{tab:certificate-tau}
\input{tables/new_certificate_tau}
\end{table}
Table~\ref{tab:certificate} reports two quantities at each level $\alpha$: the \emph{coverage}, the fraction of inputs the stability check accepts, and the \emph{violation rate}, the fraction of accepted inputs whose top-1 changed nonetheless. Validity requires the second to fall below $\alpha$, and usefulness requires the first to be large. The violation rate runs at least six times below $\alpha$ in every classification row, and no accepted retrieval query had its top-1 change at any level. Evaluating the top-1 part of Lemma~\ref{lem:cuts} against an input's own $\varepsilon$ would instead hold identically, and so test nothing.

\paragraph{Comparison with confidence scores.} Split conformal calibration makes \emph{any} score valid, so the fact that our stability check respects its bound says nothing about the theory. What separates scores is how many inputs each accepts at the same guarantee. Table~\ref{tab:baselines} runs maximum softmax probability, the standard confidence score, through the same protocol, under the generic construction that makes any score valid (threshold at the $(1-\alpha)$ quantile of the score among flipped calibration inputs).

At $\alpha = 0.10$, our stability check accepts $85.2$\% against maximum softmax probability's $74.4$\%, $80.0$ against $69.4$ at $\alpha=0.05$, and $67.7$ against $61.1$ at $\alpha=0.01$, with violation rates far inside the level in every row. The margin itself, under the generic construction, sits between the two ($82.3$, $77.9$ and $69.1$\%) and matches ours at the strictest level. So the gain over \textsc{msp} comes from the quantity being measured, distance to the decision boundary rather than confidence. The stability check of \S\ref{par:certificate} adds a little more by calibrating the perturbation of the score differences rather than a generic quantile.

\begin{table}[t]
\centering
\footnotesize
\caption{\textbf{Proposition~\ref{prop:pflip} predicts the shape as well as the level.} Proposition~\ref{prop:pflip} models the perturbation as uniform on $[-\varepsilon, \varepsilon]$. Real perturbations are not uniform, so we fit one interval half-width per group, the \emph{effective perturbation scale} $s$, reported as a fraction of $\varepsilon$ and chosen so the predicted \emph{aggregate} flip rate matches the observed one. The columns for each $|\cset|$ are then predictions with no further tuning. For classification $s \approx 0.9$: the worst-case $\varepsilon$ is already close to the operative scale. For retrieval $s$ is $0.12$--$0.24$, the inflation from taking a maximum over tens of thousands of candidates.}
\label{tab:pflip}
\input{tables/new_pflip}
\end{table}
\paragraph{The predicted curve.} Proposition~\ref{prop:pflip} turns the gap profile into a
predicted flip probability. Table~\ref{tab:pflip} fits one effective-scale parameter per group on
the aggregate rate and reports the columns for each $|\cset|$ as predictions. The shape follows in
both task families: for classification the fitted scale is $0.92$ and $0.87$, so the worst-case $\varepsilon$ is already close to the perturbation scale that matters, and the prediction needs almost no tuning. For retrieval it is $0.12$--$0.24$. This reflects the fact that $\varepsilon$ is a maximum over tens of thousands of candidates, so the scale is not a free parameter.

\paragraph{Scope.} The relation between $|\cset|$ and the change rate holds at W4, the regime in which the stability check is
useful. It does \emph{not} hold at W3, where retrieval remains $40$--$50$ points more fragile than classification at every matched width. When the perturbation is as large as the score range itself, the number of contenders no longer limits the flips. The direction of the perturbation starts to matter, and the width does not capture that. This is not an artifact of the threshold. A valid bound must take the maximum over all candidates of that input, and the candidate set differs between classification and retrieval, so no single threshold can be both valid and comparable across the two. We therefore state the guarantee with the worst-case
$\varepsilon = \lVert z^{\ptq}-z^{\fp}\rVert_\infty$ and report that relation per
bit-width.

\subsection{The \textsc{clip} results in full}\label{app:clip}

Every retrieval measurement reported on the text arm is repeated here on \textsc{clip}
text-to-image retrieval (Flickr30k, ViT-B/32 and ViT-L/14, zero-shot), from the same stored runs
as Tables~\ref{tab:clip} and~\ref{tab:alloc-clip}. Each table names its text counterpart.

\begin{table}[t]
\centering
\footnotesize

\caption{\textbf{Contender set and separation on \textsc{clip}.} Both encoders at group\_128, \rtn{}: top-1 change rate, median separation ratio $\gapk{2}/2\varepsilon$, fraction of queries whose contender set is empty, fraction of changed top-1 results taken by the full-precision runner-up, and fraction of queries whose top-5 set is unchanged. Counterpart of Figures~\ref{fig:sephist} and~\ref{fig:law}.}
\label{tab:clip-mechanism}
\input{tables/new_clip_mechanism}
\end{table}

\paragraph{Mechanism.} At W4 the contender set is empty for $1.1$\% of ViT-L/14 queries and for
none of ViT-B/32, against $74.3$\% of classification inputs (\S\ref{sec:mechanism}). The
\fp{} runner-up takes the vacated first place in $49$\% and $33$\% of the changed results,
the rest coming from deeper in the ranking, and the exact top-5 set survives for $12$\% and $3$\%
of queries. The two encoders order by separation, as the six text embedders do: ViT-L/14 at a
median ratio of $0.151$ changes $24.6$\% of its top-1 results, ViT-B/32 at $0.079$ changes $47.0$\%.
The $30$ \textsc{clip} configurations, under six quantizers and four corpus sizes, are the starred
points of Figure~\ref{fig:law}, and they sit on the same curve as the other $1{,}098$.

\paragraph{Stability check.} Under the protocol of \S\ref{par:certificate}, the check accepts $15.7$\% of queries on ViT-L/14
at $\alpha = 0.10$ and $1.5$\% on ViT-B/32, with no violation in any split,
alongside $11.3$\% on text retrieval and $85.2$\% on classification (Table~\ref{tab:certificate}).

\paragraph{What a changed result costs.} Flickr30k pairs each caption with one image, so a
changed top-1 either loses the correct image, gains it, or moves between two wrong ones. Pooled over all changed results at W4, $26$\% of ViT-L/14's and $30$\% of ViT-B/32's lose the correct image, and $23$\% and $13$\% gain it. On \textsc{beir} under the same pooling the figures are $14.6$\% and $11.5$\% (Table~\ref{tab:harm}), but there two thirds of the changes move between unjudged documents, and no such unjudged mass exists on Flickr30k. At W3 the losing share rises to $45$\% and $48$\%. Losses
and gains partly cancel in the mean, which is why Recall@1 moves by a point while a quarter of the
results change (\S\ref{sec:systems}).

\begin{table}[t]
\centering
\footnotesize

\caption{\textbf{Separation and change rate on \textsc{clip} under six quantizers.} Median separation ratio $\gapk{2}/2\varepsilon$, top-1 change rate and Recall@1 at group\_128, both encoders quantized, all $2{,}000$ queries. Counterpart of Table~\ref{tab:regime}.}
\label{tab:regime-clip}
\input{tables/new_clip_regime}
\end{table}

\paragraph{The five other quantizers.} \gptq{}, with or without activation ordering, removes a third to a half of the change rate
on both encoders at both bit-widths, AdaRound three tenths to two fifths, and HQQ and \awq{} a tenth to a third, the same ordering as on text
(Tables~\ref{tab:quantizers} and~\ref{tab:regime}). On ViT-L/14 at W4, Recall@1 under \gptq{} is
a point above \fp{} while $16.0$\% of its top-1 results differ from it. All five
raise the separation, \gptq{} with activation ordering the most, and the largest value any of them reaches is $0.278$, nearly four
times below the stability threshold.

\begin{table}[t]
\centering
\footnotesize

\caption{\textbf{Corpus size on \textsc{clip}.} Top-1 change rate at the smallest and largest image index, \rtn{} W4-group\_128, paired over the same $2{,}000$ queries ($^{*}$: $p<0.05$, McNemar's test), with the median separation ratio at each size. Nested seeded subsets of the index, as in Table~\ref{tab:corpus}.}
\label{tab:corpus-clip}
\input{tables/new_clip_corpus}
\end{table}

\paragraph{Corpus size.} On ViT-B/32 the change rate is flat from $500$ to $4{,}000$ images
($47.4$\% to $47.0$\%, $p = 0.84$). On ViT-L/14 it falls from $34.2$\% to $24.6$\% as the index
grows and the separation rises from $0.127$ to $0.151$, the reverse of the ViT self-retrieval in
Table~\ref{tab:corpus}. The reason is which image is on top. At the full index, the queries whose
correct image is ranked first ($59.8$\% of them on ViT-L/14) have a median separation of $0.252$
against $0.065$ for the rest, and change their top-1 for $10.9$\% against $45.1$\%. On ViT-B/32
the figures are $0.150$ against $0.044$ and $28.4$\% against $65.8$\%. At $500$ images, the
correct image is in the index for $13$\% of queries, so nearly every query ranks among wrong
images, where the top-1 is a near-tie. Contrastive training separated the matching pair from the
rest, not one wrong image from another. Quantizing only the query encoder, with the index embedded
at \fp{}, helps more here than on text, from $47.0$\% to $27.1$\% on ViT-B/32 and from
$24.6$\% to $15.3$\% on ViT-L/14, and still leaves a sixth to a quarter of the results changing.

\begin{table}[t]
\centering
\footnotesize

\caption{\textbf{Gap-driven allocation on \textsc{clip} under each quantizer.} Fraction of the W3$\to$W4 benefit captured at a 3.5-bit average, pooled over both encoders, on the three metrics of Table~\ref{tab:alloc-absolute}.}
\label{tab:alloc-main-clip}
\input{tables/new_clip_alloc_main}
\end{table}
\begin{table}[t]
\centering
\footnotesize

\caption{\textbf{Absolute \textsc{clip} retrieval metrics under each allocation.} Mean over both encoders, \rtn{}, average 3.5 bits for the middle rows. Columns as in Table~\ref{tab:alloc-absolute}.}
\label{tab:alloc-absolute-clip}
\input{tables/new_clip_alloc_absolute}
\end{table}
\begin{table}[t]
\centering
\footnotesize

\caption{\textbf{Per encoder on \textsc{clip}.} Columns as in Table~\ref{tab:alloc-permodel}: capture of the W3$\to$W4 flip-rate benefit at a 3.5-bit average under \rtn{}, and the layer-level correlation between reconstruction error and gap sensitivity.}
\label{tab:alloc-permodel-clip}
\input{tables/new_clip_alloc_permodel}
\end{table}

\paragraph{Allocation.} Gap sensitivity captures $60.9$\% of the flip-rate benefit under \rtn{}
against $37.8$\% for reconstruction error, and leads on all three metrics under every quantizer
(Table~\ref{tab:alloc-main-clip}). In absolute terms, it changes $51.2$\% of top-1 results against
$60.2$\%, reaches $43.9$\% Recall@1 against $37.0$\%, and loses $33.9$\% of correct images against
$45.6$\% (Table~\ref{tab:alloc-absolute-clip}). It wins on each encoder separately: $57.3$\% against
$25.8$\% on ViT-B/32 and $65.6$\% against $53.9$\% on ViT-L/14 (Table~\ref{tab:alloc-permodel-clip}).
Reconstruction error correlates with gap sensitivity across layers at $+0.12$ on ViT-B/32 and $+0.34$ on ViT-L/14 (Table~\ref{tab:alloc-permodel-clip}), higher than on the text embedders. The encoder with the higher correlation is also the one on which reconstruction error recovers more, $53.9$\% against $25.8$\%. On both, it stays behind gap sensitivity. The allocation is also stable across budgets and calibration seeds on
\textsc{clip} (Tables~\ref{tab:alloc-budget} and~\ref{tab:alloc-seed}).

\subsection{Allocation criteria}\label{app:alloc-criteria}

Capture is pooled as $\sum \text{gain} / \sum \text{headroom}$ rather than averaged over per-configuration ratios, so that a configuration with almost nothing to gain does not dominate through a noisy ratio. The paragraphs
below give the score and where it comes from. Every criterion scores each linear layer $\ell$
with a single number $s_\ell$. Layers are then
ranked by $s_\ell$ divided by their parameter count, since bit-width is paid per weight, and the
wider bit-width goes to the highest-ranked layers until the budget is spent. Write $W_\ell$ for the layer's weight matrix, $\Delta_\ell = W_\ell - Q(W_\ell)$ for the change quantization makes to it, and $x$ for the layer's input. Expectations over $x$ are taken on the
calibration set, in one forward pass that records, for each input channel $i$, the two moments
$\mathbb{E}[x_i^2]$ and $\mathbb{E}|x_i|$.

\paragraph{Gap sensitivity (ours).} Quantize layer $\ell$ alone, leave the rest at \fp{},
and record how far the \fp{} top-1/top-2 gap moves, as in
\S\ref{sec:allocation}. It is the only criterion that scores a layer by its effect on the quantity
the ranking depends on rather than on an error norm. Needs calibration queries.

\paragraph{Reconstruction error.} $s_\ell = \lVert \Delta_\ell \rVert^2 / |W_\ell|$, the mean
squared change to the layer's weights. This is the objective weight-space allocators minimize: per-channel bit allocation at a fixed average bit-width solves exactly this problem~\citep{banner2019post}, and it is the weight-space factor that HAWQ-v2 weights by the Hessian trace~\citep{dong2020hawqv2}. It is the main text's baseline. Data-free.

\paragraph{Relative error (SQNR).} $s_\ell = \lVert \Delta_\ell \rVert^2 / \lVert W_\ell \rVert^2$,
the same quantity made scale-free, so a layer with large weights is not ranked highly merely for
having large weights. Per-layer SQNR of this form is the classical basis for allocating fixed-point bit-widths across layers~\citep{lin2016fixed}, and SQNR remains the sensitivity signal in recent post-training mixed-precision allocators~\citep{pandey2023practical}, which measure it at the network output rather than on the weights. Data-free.

\paragraph{Activation-weighted error.} $s_\ell = \sum_i \mathbb{E}[x_i^2] \,
\lVert \Delta_\ell[:,i] \rVert^2$, reconstruction error measured in activation space rather than
weight space. It is the diagonal approximation of the layerwise proxy loss \gptq{}~\citep{frantar2023gptq} minimizes, since that layer Hessian is $2X^\top X$ and its diagonal is proportional to $\mathbb{E}[x_i^2]$. The proxy predates \gptq{}: \citet{nagel2020adaround} derive it from a second-order expansion of the task loss, and \citet{frantar2022obc} solve it in Optimal Brain Surgeon form. It needs calibration data and no backward pass.

\paragraph{Activation salience.} $s_\ell = \sum_i \mathbb{E}|x_i| \,
\lVert \Delta_\ell[:,i] \rVert^2$, the same form with the first absolute moment rather than the
second. \awq{}~\citep{lin2024awq} protects channels with large activations by rescaling
them. This criterion applies the premise per layer rather than per channel. Needs
calibration data.

\paragraph{Activation norm.} $s_\ell = \sum_i \mathbb{E}[x_i^2]$, the energy passing through the
layer, ignoring what quantization does to its weights. We are not aware of prior work that ranks layers this way. It is included as a control to separate data-awareness from the rest, and tests whether knowing which layers see large activations is by itself enough to rank them. Needs calibration data.

\paragraph{Depth heuristic.} $s_\ell = |d_\ell - \bar{d}| / \tfrac{1}{2}(d_{\max} - d_{\min})$,
the layer's distance from the middle of the stack, so the first and last blocks rank highest. This
is inspired by the common practice of keeping the first and last layers of a quantized network at higher precision, reported as more sensitive to quantization~\citep{zhou2016dorefa,choi2018pact,wu2020integer}. Data-free and weight-free.

Three of the six baselines therefore depend on the calibration set exactly as gap sensitivity
does. In the calibration-size ablation of Table~\ref{tab:alloc-calib}, every criterion is recomputed
from the same calibration slice at each size, so a comparison at a small budget does not silently
give the baselines a larger one.

\subsection{Allocation robustness: budgets and calibration seeds}\label{app:alloc-robust}
\begin{table}[t]
\centering
\footnotesize

\caption{\textbf{The allocation holds across budgets.} Capture of the W3$\to$W4 benefit at three fractional bit budgets under \rtn{}, calibration seed 2038, pooled over six text embedders and four corpora (Text) and over both \textsc{clip} encoders. At an integer average uniform is the only allocation, so the question is only posed between integers.}
\label{tab:alloc-budget}
\input{tables/new_alloc_budget}
\end{table}
\paragraph{Deployment.} Every layer is entirely at 3 or at 4 bits, so the allocation needs no new low-level code: each layer runs with the ordinary 3-bit or 4-bit matrix-multiply routine, dispatched per layer.

\paragraph{Budget generality.} The result does not depend on the $3.5$-bit budget (Table~\ref{tab:alloc-budget}): gap sensitivity is first at $3.25$ and $3.75$ bits per weight as well.
Capture rises with the budget, as it must, since there are more bits to place, and
the ordering is unchanged at every budget.

\begin{table}[t]
\centering
\footnotesize

\caption{\textbf{The allocation is stable across calibration draws (\rtn{}).} Fraction of layers receiving the same bit-width under different calibration seeds $\{101,202,2038\}$, over all corpora and budgets on the text arm and at a $3.5$-bit average on \textsc{clip}. Chance agreement for this two-way split is approximately $50$\%.}
\label{tab:alloc-seed}
\input{tables/new_alloc_seed}
\end{table}

\paragraph{The allocation is stable.} A sensitivity estimated from 128 calibration queries could be fitting that particular sample (Table~\ref{tab:alloc-seed}). It is not: across 72 seed pairs, two calibration samples
agree on the bit-width of $91.4\%$ of layers, against roughly $50\%$ by chance for a two-way split. The layers that decide a ranking are a property of the model, not of the sample used to
find them.

\begin{table}[t]
\centering
\footnotesize
\setlength{\tabcolsep}{3pt}
\caption{\textbf{Gap sensitivity leads at every calibration size.} $3.5$-bit average under \rtn{}. Text rows pool six embedders on four \textsc{beir} corpora, $24$ runs each, and \textsc{clip} rows are one run per cell. \emph{Agreement} is the fraction of layers given the same bit-width as the $128$-query gap allocation of the same system, with the query pool fixed so only the sample size varies. Chance for this two-way split is about $50$\%, and on text two $128$-query allocations drawn from different pools agree on $93.2$\%. \emph{Capture} is the share of the W3$\to$W4 reduction in top-1 change recovered by each criterion, pooled as $\sum$gain$/\sum$headroom, every criterion recomputed from the same calibration slice. The best per row is in bold. On text, gap sensitivity beats all six baselines in $13$ of $24$ runs at $8$ queries and in $18$ to $22$ of $24$ at larger sizes. \textsc{scidocs} has $1000$ queries and the calibration slice is capped at a quarter of the pool, so its largest arm uses $250$ rather than $256$.}
\label{tab:alloc-calib}
\input{tables/new_alloc_calib}
\end{table}

\paragraph{Calibration size.} Agreement with the $128$-query allocation (Table~\ref{tab:alloc-calib}) rises from $80.6$\% at $8$ queries to $95.2$\% at $256$, against $93.2$\% between two $128$-query draws from different pools, so the allocation has converged by $64$ queries. Capture rises with the sample, from $66.6$\% at $8$ queries to $73.8$\% at $128$, and is flat beyond. The three data-free criteria stay at $45$--$47$\% at every size, as they must, and the activation-based ones do not improve with more data. Gap sensitivity beats all six baselines in $13$ of $24$ runs at $8$ queries and in $20$ to $22$ of $24$ from $128$. On \textsc{clip} the ordering is the same at every size on both encoders (Table~\ref{tab:alloc-calib}).

\subsection{Activation quantization}\label{app:activation}
Every result in the main text quantizes the weights only. Here we also quantize the activations to eight bits at the input of every linear layer. Each activation tensor gets one fixed scale. We set the scale before deployment by running the model on the same calibration data the weight quantizers use, recording the largest value the tensor takes, and keeping that value fixed from then on.

\textsc{clip} is excluded because our method cannot reach all of its activations. We quantize an activation by attaching a small function to a layer that intercepts the layer's input just before the layer runs. This works whenever the layer is called in the normal way. In \textsc{clip}, the attention output projection is not called in the normal way. The attention implementation in PyTorch applies that layer's weights directly inside one combined function, so the layer itself is never called and our function is never reached. This affects $24$ of the model's $72$ linear layers. We could still quantize the weights of those layers, since weights are changed in place, but not their activations. A run would then quantize activations on two-thirds of the layers and leave the rest untouched, and reporting that as activation quantization would overstate what was done. Every other model in the paper calls all of its linear layers in the normal way, so all of them are covered.

Table~\ref{tab:activation} reports the $165$ configurations measured at W4, W4A8 and W8A8, at group\_128 under \rtn{}. Three things hold. The asymmetry is unchanged: retrieval changes $6.6\times$ more top-1 results than classification at W4, $6.9\times$ at W4A8, and $8.1\times$ at W8A8. Among the activation-quantized configurations alone, the separation ratio still predicts the change rate, at Spearman $-0.90$ over all $168$ configurations at W4 and $-0.93$ over $165$ at W8. The relation also carries over from one source of perturbation to the other: a line fitted on the weight-only configurations at four and three bits predicts the activation-quantized ones it never saw at $R^2 = 0.88$ in both cases, with the same slope to within $0.02$. The perturbation added by eight-bit activations is comparable in size to the error from W4 and much larger than the error from W8, which is why W8A8 damages retrieval more than W4 alone.

\begin{table}[t]
\centering
\footnotesize
\caption{\textbf{Eight-bit activations on top of W4 and W8.} Top-1 change rate, median separation ratio, and the fraction of configurations above the stability threshold, over the $165$ configurations present in all three settings, group\_128 under \rtn{}. Every column is measured against \fp{}.}
\label{tab:activation}
\input{tables/new_activation}
\end{table}

The result does not depend on the weight quantizer. Table~\ref{tab:activation-quantizers} repeats the comparison under each of the six weight quantizers, at W4 and at W8, with the activation scheme unchanged. Under every quantizer, adding eight-bit activations raises the classification change rate to between $7$ and $9$\% at W4 and to $6$ or $7$\% at W8, and the retrieval change rate to between $56$ and $60$\% and to $52$ or $53$\%. Hence, retrieval changes seven to eight times more top-1 results than classification in every case. The separation ratio predicts the change rate over the activation-quantized configurations alone at Spearman $-0.90$ or stronger under every quantizer and both bit-widths. The weight quantizer matters little once activations are quantized: at W4 alone the retrieval change rate spans $20$ to $30$\% across quantizers, but with activations added, it spans $56$ to $60$\%, and at W8, the six quantizers agree to within half a point. The activation error is then the larger of the two, and the error-minimizing weight quantizers lose their advantage.

\begin{table}[t]
\centering
\footnotesize
\caption{\textbf{Eight-bit activations under every weight quantizer.} Top-1 change rate at W4 and W8, without and with eight-bit activations (W4A8, W8A8), for classification and retrieval, the retrieval-to-classification ratio with activations quantized, and Spearman between median separation ratio and change rate over the activation-quantized configurations alone. Over the same $165$ configurations as Table~\ref{tab:activation}, group\_128.}
\label{tab:activation-quantizers}
\resizebox{\linewidth}{!}{\input{tables/new_activation_quantizers}}
\end{table}

\FloatBarrier
Our gap-sensitivity allocation also survives activation quantization under every weight quantizer. We re-evaluate the stored allocations of the $24$ text retrieval configurations, six embedders on four \textsc{beir} corpora, with eight-bit activation quantization, under every policy, so the allocation still assigns weight bits and the activation error is a fixed extra perturbation. Table~\ref{tab:alloc-activation} gives the pooled capture under each of the six weight quantizers, with and without activations. Gap sensitivity is first under all six, capturing $60$ to $76$\% of the extra bit against $33$ to $53$\% for reconstruction error, and it is first per configuration in $17$ to $24$ of the $24$. Under \rtn{}, HQQ, \gptq{} with activation ordering and AdaRound, reconstruction error again does not outperform a random split. The closest pair is \gptq{}, where activation quantization lifts reconstruction error to $53$\% against $60$\% for gap sensitivity.

\begin{table}[t]
\centering
\footnotesize
\caption{\textbf{Gap allocation still leads when activations are quantized too, under every weight quantizer.} Share of the W3$\to$W4 reduction in top-1 change recovered at a $3.5$-bit average, pooled over the $24$ text retrieval configurations, six embedders on four \textsc{beir} corpora, without (W) and with eight-bit activations (W+A8).}
\label{tab:alloc-activation}
\input{tables/new_alloc_activation_quantizers}
\end{table}

\section{Theory: deferred statements and remarks}\label{app:theory-detail}

\paragraph{Why indifference is not transitive.} Two candidates are indifferent when neither
is certainly above the other. Certain dominance itself is transitive, because gaps add: if
$z_i - z_j \ge 2\varepsilon$ and $z_j - z_k \ge 2\varepsilon$ then $z_i - z_k \ge 4\varepsilon$.
Indifference is not, because a chain of candidates each within $2\varepsilon$ of the next can span
an arbitrarily large total gap. This is why a candidate sitting far below the top-1 can still
reach the first position, which \S\ref{sec:mechanism} measures.

\paragraph{Boundary cases and verification.} The definition of certain dominance uses $\ge$ rather
than $>$, and the choice matters at the boundary. When $z_i - z_j$ is exactly $2\varepsilon$, no
perturbation of size $\varepsilon$ can put $j$ strictly above $i$, because the most it can do is
bring the two level. The pair is therefore genuinely ordered and belongs in the relation. The proof
of Proposition~\ref{prop:reachable} rests on the equivalent statement that a ranking is realizable iff
$z_{c_k} - z_{c_j} < 2\varepsilon$ for all positions $j < k$. We also checked it by
exhaustive enumeration over $1.3$M permutations in exact rational arithmetic, covering both generic
score vectors and ones with gaps sitting exactly at $2\varepsilon$, with no mismatches.

\paragraph{Why each score is limited separately.} Proposition~\ref{prop:reachable} assumes that no
single score moves by more than $\varepsilon$, which is what $\lVert \delta \rVert_\infty \le
\varepsilon$ says, and the characterization depends on the limit having that form. Suppose instead
the limit applied to the perturbation as a whole, as $\lVert \delta \rVert_2 \le \varepsilon$
does. Concentrating $\delta$ on a single score then reaches rankings that spreading it across
many cannot, so whether a ranking is reachable depends on how many pairs it reverses at once, and
no single threshold on a pair can describe the reachable set. Weight quantization rounds each weight
on its own, and the error it puts on each score is bounded separately, so the per-score limit
matches what quantization does.

\paragraph{Why the ratio and not the gap.} The scores in the two task families are not measured in
the same units. A logit is unbounded and a cosine similarity lies in $[-1,1]$, so a reader may
worry that comparing $\varepsilon$ across the two settings compares nothing. We never do. Every
comparison is on the separation ratio $\gapk{2}/2\varepsilon$, which is dimensionless and
invariant under any positive affine change of the scores. Replacing $z$ by $cz + b$ with $c > 0$
scales both $\gapk{2}$ and $\varepsilon$ by $c$ and cancels $b$ in both, so the ratio is unchanged,
and so is the ranking and therefore every flip we count. Temperature scaling of logits is the case
$c = 1/\tau$. The invariance does not survive a nonlinear map: on $z = (3,2,1)$ with
$\delta = (0, 0.4, 0)$, the ratio is $1.25$ on the logits and $2.59$ after a softmax, which is a
further reason to measure the gap on the scores a system ranks by rather than on probabilities.

\paragraph{The general characterization.} The whole ranking survives every perturbation exactly when $\mathcal{R}_\varepsilon(z) = \{\pi(z)\}$, and Lemma~\ref{lem:cuts} is the case of the first position and of the first $k$.

\begin{proposition}[Characterization of reachable rankings]\label{prop:reachable}
$\sigma = (c_1, \dots, c_N) \in \mathcal{R}_\varepsilon(z)$ iff $z_{c_k} - z_{c_j} < 2\varepsilon$ for all positions $j < k$.
\end{proposition}

Both parts of Lemma~\ref{lem:cuts} follow by applying this to one pair. We use only the top-1 part, but the general statement is about a whole ranking rather than one comparison, which is what the non-transitivity above and \S\ref{sec:mechanism} rest on.

\paragraph{From reachability to probability.} Proposition~\ref{prop:reachable} says which rankings are
reachable, not how often one is realized, and on retrieval the resulting bound says almost nothing.
Almost every input has a nonempty contender set, so the statement that the ranking may change is
true and useless. The gap profile holds more information than the bound uses, and a short
calculation extracts it.

\begin{proposition}[Flip probability]\label{prop:pflip}
Let $\delta_1, \dots, \delta_N$ be i.i.d.\ uniform on $[-\varepsilon, \varepsilon]$ and write the separation ratio
$a_j = \gapk{j} / 2\varepsilon$. Then the top-1 survives with probability
\[
  \Pr[\text{top-1 stable}] \;=\; \int_0^1 \prod_{j \in \cset} \min(u + a_j,\, 1)\; du .
\]
\end{proposition}

Two extreme cases check the formula. Write $w = |\cset|$ for the number of contenders. If every
contender is tied with the winner, so that $a_j = 0$ for all of them, the integral becomes
$\int_0^1 u^{\,w} du = 1/(w+1)$. That is what you would predict by treating the winner and its $w$
contenders as interchangeable and asking for the chance that the winner comes out on top, so the
formula agrees with the naive count in the case where the naive count should apply. At the other
extreme, if every contender sits exactly at the boundary, so that $a_j = 1$, the integral is $1$ and
the flip probability is zero, which is Lemma~\ref{lem:cuts} again. Real inputs fall between the
two, and the interchangeable estimate systematically predicts too many flips there, because
contenders sit at different distances from the winner rather than all at zero.
Proposition~\ref{prop:pflip} accounts for those distances.

Table~\ref{tab:pflip} evaluates the prediction. Proposition~\ref{prop:pflip} assumes the
perturbations are independent and uniform, which is a modeling choice rather than something
quantization guarantees. Whether it captures the shape of the measured curve is an empirical
question.

\paragraph{Which score to route on.} The routing rule of \S\ref{sec:routing} thresholds the quantized gap. The following says that, up to order-equivalence, no other score matches the flippable sets at every budget.

\begin{proposition}[Identifiability of the gap]\label{prop:routing}
Let $\mathcal S$ be the class of measurable scores $s(x) = g(z_{\mathrm{q}}(x))$ of the quantized
logit vector $z_{\mathrm{q}}(x)$, and assume the gap \qmargin{} has an atom-free distribution. Then $s$ admits, for every
$\varepsilon > 0$, a threshold $\theta_\varepsilon$ such that $\{x : s(x) \le \theta_\varepsilon\}$ equals the
$\varepsilon$-flippable set up to a null set, if and only if $s$ preserves the strict order of \qmargin{} on almost every pair of
inputs. In particular \qmargin{} itself qualifies, with threshold $2\varepsilon$, and any score
that re-ranks a positive-measure set of pairs against it fails at some budget.
\end{proposition}

This is an identifiability statement, not an optimality one: once Lemma~\ref{lem:cuts} fixes the flippable sets, only order-equivalents of the gap can match them at every budget. It concerns the worst-case flippable set at each budget, not which inputs a given perturbation actually changes. The rule of \S\ref{sec:routing} uses one threshold rather than every budget, so the support for using the gap there is empirical, not this proposition. Proof in Appendix~\ref{app:proofs}.

\section{Proofs}\label{app:proofs}
Full proofs of the statements in Section~\ref{sec:theory} and Appendix~\ref{app:theory-detail}.

\paragraph{Proof of Proposition~\ref{prop:reachable}.} Write $P_\varepsilon$ for the relation $i \prec j$ iff $z_i - z_j \ge 2\varepsilon$, and call a permutation a linear extension of $P_\varepsilon$ if it places $i$ before $j$ whenever $i \prec j$.
\emph{Necessity.} Suppose $\delta$ realizes $\pi$ and $i \prec j$ in $P_\varepsilon$, i.e.
$z_i - z_j \ge 2\varepsilon$. Note $i \ne j$, since $\varepsilon > 0$ makes $P_\varepsilon$
irreflexive. Then
\[
  (z_i + \delta_i) - (z_j + \delta_j) \;\ge\; (z_i - z_j) - |\delta_i| - |\delta_j| \;\ge\; 0 .
\]
The perturbed values are strictly ordered along $\pi$, hence pairwise distinct, so the inequality
is strict and $i$ precedes $j$. Thus $\pi$ is a linear extension.

\emph{Sufficiency.} Let $\pi$ be a linear extension. For $j < k$, if
$z_{c_k} - z_{c_j} \ge 2\varepsilon$ then $c_k \prec c_j$ in $P_\varepsilon$ and $\pi$ would have
to place $c_k$ first, a contradiction, so $z_{c_k} - z_{c_j} < 2\varepsilon$ for all $j < k$. Set
\[
  t_1 = z_{c_1} + \varepsilon, \qquad
  t_k = \min\!\left( z_{c_k} + \varepsilon,\; t_{k-1} - \eta \right),
\]
which unrolls to $t_k = \min_{j \le k} \left( z_{c_j} + \varepsilon - (k-j)\eta \right)$. Then $t$
is strictly decreasing and $t_k \le z_{c_k} + \varepsilon$ by construction. The remaining
requirement $t_k \ge z_{c_k} - \varepsilon$ holds provided
$z_{c_k} - z_{c_j} \le 2\varepsilon - (k-j)\eta$ for all $j \le k$. The instance $j = k$ reads
$0 \le 2\varepsilon$ and is trivial. Choosing
\[
  0 \;<\; \eta \;<\; \min_{j < k} \frac{2\varepsilon - (z_{c_k} - z_{c_j})}{k - j},
\]
a minimum over a nonempty finite set of strictly positive reals, makes this hold. Then
$\delta_{c_k} = t_k - z_{c_k}$ satisfies $\lVert \delta \rVert_\infty \le \varepsilon$ and
realizes $\pi$.

\paragraph{Proof of Lemma~\ref{lem:cuts}.}
\emph{Top-1.} Apply Proposition~\ref{prop:reachable} with the pair $(z_{(1)}, z_{(2)})$: if
$z_{(1)} - z_{(2)} \ge 2\varepsilon$, then no linear extension places any other candidate first, so
every realizable ranking has the same leader. Conversely, if the gap is below $2\varepsilon$, the
transposition swapping the top two is a linear extension, hence realizable by Proposition~\ref{prop:reachable}.

\emph{Top-$k$.} If $z_{(k)} - z_{(k+1)} \ge 2\varepsilon$ then every candidate at rank $\le k$ certainly dominates
every candidate at rank $> k$, so by Proposition~\ref{prop:reachable} no realizable ranking moves one
across the cut and the top-$k$ set is invariant. If the gap is below $2\varepsilon$, ranks $k$ and
$k+1$ are incomparable, so the transposition exchanging them extends to a linear extension and
changes the set.

\paragraph{Proof of Corollary~\ref{cor:local}.}
Candidate $j$ overtakes $a$ exactly when $z_a + \delta_a < z_j + \delta_j$, i.e.\ when $\delta_j - \delta_a > z_a - z_j$. The top-1 survives iff this fails for every $j$, an event that depends on $\delta$ only through the differences $\delta_j - \delta_a$. For the second claim suppose the top-1 changed, so $a' \ne a$ with $z_a > z_{a'}$ and $z_{a'} + \delta_{a'} \ge z_a + \delta_a$. Let $b'$ be the runner-up of $z + \delta$. Since $z_a + \delta_a \le z_{b'} + \delta_{b'}$,
\[
  \gapk{2}(z + \delta) = (z_{a'} + \delta_{a'}) - (z_{b'} + \delta_{b'}) \le (z_{a'} - z_a) + (\delta_{a'} - \delta_a) < \delta_{a'} - \delta_a \le S,
\]
using $z_{a'} - z_a < 0$. Finally, $\delta_{a'} - \delta_j \le 2\lVert \delta \rVert_\infty$ for every $j$, so $S \le 2\lVert \delta \rVert_\infty$, with equality only when two coordinates are perturbed by $\varepsilon$ in opposite directions. Calibrating $S$ directly avoids that loss. The single pair of top-1 and runner-up does \emph{not} suffice: a third candidate can overtake $a$ while $z_a - z_b$ is unchanged, which is why $S$ maximizes over all challengers.

\paragraph{Proof of Proposition~\ref{prop:routing}.}
\emph{Necessity.} Suppose $\qmargin(x) < \qmargin(x')$. Pick a rational
$c \in (\qmargin(x), \qmargin(x'))$ and apply the hypothesis at
$\varepsilon = c/2$: there is a threshold $\theta$ with
$\{z : s(z) \le \theta\} = \{z : \qmargin(z) < c\}$ up to a null set $N_c$. Outside $N_c$,
$x$ lies in the right-hand set so $s(x) \le \theta$, and $x'$ does not, so $s(x') > \theta$.
Hence $s(x) < s(x')$. The union of exception sets over rational $c$ is null.

\emph{Sufficiency.} Fix $\varepsilon > 0$ and $c = 2\varepsilon$, and let
$A = \{\qmargin < c\}$, $B = \{\qmargin > c\}$, and $\theta = \operatorname*{ess\,sup}_A s$.
Then $A \subseteq \{s \le \theta\}$ up to a null set. If
$B_0 = \{x' \in B : s(x') \le \theta\}$ had positive measure, order preservation would force
$s = \theta$ almost everywhere on $B_0$, hence (applying order preservation within
$B_0 \times B_0$) \qmargin{} constant on $B_0$ up to a null set, contradicting atom-freeness.
Since $\{\qmargin = c\}$ is null by atom-freeness, $\{s \le \theta\}$ equals the
$\varepsilon$-flippable set up to a null set. The flippable set itself equals
$\{\qmargin < 2\varepsilon\}$ by Lemma~\ref{lem:cuts} and its matching converse
(the two-coordinate perturbation $z_i \mapsto z_i - \varepsilon$,
$z_j \mapsto z_j + \varepsilon$ flips any input with $\qmargin < 2\varepsilon$). \hfill$\square$

\paragraph{Remark.} Order preservation, not a pointwise factorization $s = h \circ \qmargin$,
is the exact condition threshold routing needs: it makes the bottom-$p$ set of $s$ coincide
with the bottom-$p$ set of \qmargin{} for every budget $p$, up to null sets.

\paragraph{Proof of Proposition~\ref{prop:pflip}.}
The top-1 is displaced by $j$ iff $\delta_j - \delta_1 \ge \gapk{j}$, so it survives iff
$\delta_j < \delta_1 + \gapk{j}$ for every $j$. Conditioning on $\delta_1$ and using independence,
$\Pr[\text{stable} \mid \delta_1] = \prod_j \Pr[\delta_j < \delta_1 + \gapk{j}]$. For $\delta_j$
uniform on $[-\varepsilon,\varepsilon]$, $\Pr[\delta_j < t] = \min\big((t+\varepsilon)/2\varepsilon,\, 1\big)$
truncated below at $0$. Substituting $u = (\delta_1 + \varepsilon)/2\varepsilon$, which is uniform
on $[0,1]$, gives $\Pr[\delta_j < \delta_1 + \gapk{j}] = \min(u + a_j, 1)$, and integrating over
$u$ yields the claim. Candidates outside $\cset$ have $a_j \ge 1$ and contribute a factor of one.

\end{document}

%% file: tables/new_fragility.tex
\begin{tabular}{l|cc|c}
\toprule
Backbone & Classification & Retrieval & Ratio \\
\midrule
ViT-B/16 & 6.2\% \;\scriptsize$\pm$1.4 \;\scriptsize($n$=21) & 45.5\% \;\scriptsize$\pm$2.5 \;\scriptsize($n$=5) & $\mathbf{7.3\times}$ \\
ViT-L/16 & 2.9\% \;\scriptsize$\pm$0.7 \;\scriptsize($n$=21) & 42.1\% \;\scriptsize$\pm$1.5 \;\scriptsize($n$=5) & $\mathbf{14.3\times}$ \\
Qwen3-Emb-0.6B & 4.3\% \;\scriptsize$\pm$1.1 \;\scriptsize($n$=11) & 33.4\% \;\scriptsize$\pm$4.1 \;\scriptsize($n$=4) & $\mathbf{7.8\times}$ \\
\midrule
Qwen3-Emb-0.6B, one checkpoint & 8.6\% \;\scriptsize$\pm$3.1 \;\scriptsize($n$=3) & 41.2\% \;\scriptsize$\pm$4.9 \;\scriptsize($n$=3) & $\mathbf{4.8\times}$ \\
\bottomrule
\end{tabular}

%% file: tables/new_gold.tex
\begin{tabular}{l|c|cccc}
\toprule
& & & \multicolumn{3}{c}{Gold document lost from top-$k$} \\
\cmidrule(lr){4-6}
Model & Bits & NDCG loss & $k{=}1$ & $k{=}5$ & $k{=}10$ \\
\midrule
Qwen3-Emb-0.6B & W4 & $-7.5$\% & 22.2\% & 23.7\% & 23.3\% \\
Qwen3-Emb-8B & W4 & $-3.1$\% & 15.6\% & 16.2\% & 15.3\% \\
BGE-large & W4 & $-2.7$\% & 16.0\% & 17.8\% & 16.3\% \\
E5-large-v2 & W4 & $-7.6$\% & 18.6\% & 21.6\% & 23.7\% \\
\midrule
Qwen3-Emb-0.6B & W3 & $-68.9$\% & 79.6\% & 80.8\% & 78.4\% \\
\bottomrule
\end{tabular}

%% file: tables/new_certificate.tex
\begin{tabular}{l|c|cc}
\toprule
Setting & $\alpha$ & Coverage & Violation \\
\midrule
Classification (ViT-B/16) & 0.10 & 85.2\% $\pm$ 0.1 & 1.47\% \\
 & 0.05 & 80.0\% $\pm$ 0.1 & 0.72\% \\
 & 0.01 & 67.7\% $\pm$ 0.2 & 0.10\% \\
\addlinespace
Text retrieval (Qwen3) & 0.10 & 11.3\% $\pm$ 0.4 & 0.00\% \\
 & 0.05 & 8.8\% $\pm$ 0.3 & 0.00\% \\
 & 0.01 & 4.8\% $\pm$ 0.4 & 0.00\% \\
\addlinespace
\textsc{clip} ViT-B/32 & 0.10 & 1.5\% $\pm$ 0.4 & 0.00\% \\
 & 0.05 & 1.1\% $\pm$ 0.3 & 0.00\% \\
 & 0.01 & 0.5\% $\pm$ 0.2 & 0.00\% \\
\addlinespace
\textsc{clip} ViT-L/14 & 0.10 & 15.7\% $\pm$ 0.9 & 0.00\% \\
 & 0.05 & 12.5\% $\pm$ 0.9 & 0.00\% \\
 & 0.01 & 7.8\% $\pm$ 0.9 & 0.00\% \\
\bottomrule
\end{tabular}

%% file: tables/new_clip.tex
\begin{tabular}{l|c|cccc}
\toprule
Encoder & Bits & Top-1 change & R@1 (\fp) & R@1 (\ptq) & Correct image lost$^{\ast}$ \\
\midrule
ViT-B/32 & W4 & 47.0\% & 50.1\% & 41.9\% \;\scriptsize($-16.4$\%) & 28.4\% \\
ViT-B/32 & W3 & 91.9\% & 50.1\% & 9.4\% \;\scriptsize($-81.2$\%) & 87.9\% \\
ViT-L/14 & W4 & 24.6\% & 59.8\% & 59.1\% \;\scriptsize($-1.3$\%) & 10.9\% \\
ViT-L/14 & W3 & 58.1\% & 59.8\% & 40.1\% \;\scriptsize($-33.0$\%) & 43.6\% \\
\bottomrule
\end{tabular}

%% file: tables/new_alloc_nograd.tex
\begin{tabular}{l|rrrrrr}
\toprule
Allocation criterion & RTN & HQQ & \gptq{} & \gptq{} (act-order) & \awq{} & AdaRound \\
\midrule
Gap sensitivity (ours) & \textbf{73.4\%} & \textbf{72.8\%} & \textbf{60.2\%} & \textbf{63.2\%} & \textbf{60.7\%} & \textbf{69.4\%} \\
\midrule
Activation norm & 55.0\% & 55.9\% & 36.8\% & 36.1\% & 40.7\% & 55.2\% \\
Reconstruction error & 47.5\% & 43.8\% & 44.2\% & 43.9\% & 46.6\% & 39.8\% \\
Relative error (SQNR) & 47.0\% & 47.7\% & 50.7\% & 51.6\% & 55.2\% & 38.9\% \\
Depth heuristic & 47.3\% & 41.3\% & 42.7\% & 43.6\% & 44.1\% & 35.8\% \\
Activation-weighted error & 44.3\% & 48.6\% & 30.5\% & 29.1\% & 38.4\% & 51.0\% \\
Activation salience & 42.8\% & 46.0\% & 30.2\% & 31.2\% & 38.7\% & 50.8\% \\
\bottomrule
\end{tabular}

%% file: tables/new_alloc_clip.tex
\begin{tabular}{l|rrr}
\toprule
Allocation criterion & RTN & \gptq{} & \awq{} \\
\midrule
Gap sensitivity (ours) & \textbf{60.9\%} & \textbf{37.9\%} & \textbf{44.8\%} \\
\midrule
Activation norm & 38.4\% & 27.5\% & 28.9\% \\
Depth heuristic & 37.6\% & 34.5\% & 27.4\% \\
Relative error (SQNR) & 39.4\% & 17.1\% & 18.8\% \\
Activation-weighted error & 38.4\% & 24.2\% & 27.9\% \\
Activation salience & 36.5\% & 33.3\% & 27.3\% \\
Reconstruction error & 37.8\% & 22.4\% & 15.6\% \\
\bottomrule
\end{tabular}

%% file: tables/new_contrast.tex
\begin{tabular}{l|l|cccc}
\toprule
Dataset & Bits & \multicolumn{2}{c}{Top-1 change rate} & \multicolumn{2}{c}{Median separation} \\
\cmidrule(lr){3-4}\cmidrule(lr){5-6}
 & & Classification & Retrieval & Classification & Retrieval \\
\midrule
Banking77 & W4 & 4.8\% & \textbf{32.0\%} & 1.16 & 0.057 \\
Banking77 & W3 & 24.6\% & \textbf{63.5\%} & 0.34 & 0.024 \\
Emotion & W4 & 14.8\% & \textbf{48.6\%} & 0.85 & 0.059 \\
Emotion & W3 & 44.2\% & \textbf{87.6\%} & 0.35 & 0.023 \\
MTOPIntent & W4 & 6.4\% & \textbf{43.1\%} & 1.39 & 0.048 \\
MTOPIntent & W3 & 28.4\% & \textbf{74.9\%} & 0.43 & 0.021 \\
\bottomrule
\end{tabular}

%% file: tables/new_arch.tex
\begin{tabular}{l|l|r|cccc}
\toprule
& & & \multicolumn{2}{c}{W4-group\_128} & \multicolumn{2}{c}{W3-group\_128} \\
\cmidrule(lr){4-5}\cmidrule(lr){6-7}
Model & Family & Params (B) & Change & Sep. & Change & Sep. \\
\midrule
Qwen3-Emb-0.6B & Decoder & 0.6 & 33.4\% & 0.060 & 86.0\% & 0.022 \\
Qwen3-Emb-4B & Decoder & 4 & 24.4\% & 0.120 & 75.2\% & 0.032 \\
Qwen3-Emb-8B & Decoder & 8 & 21.3\% & 0.143 & 46.9\% & 0.054 \\
BGE-large & Encoder & 0.34 & 22.6\% & 0.094 & 54.3\% & 0.037 \\
E5-large-v2 & Encoder & 0.34 & 29.5\% & 0.071 & 70.9\% & 0.028 \\
GTE-large & Encoder & 0.34 & 20.0\% & 0.122 & 44.8\% & 0.055 \\
\bottomrule
\end{tabular}

%% file: tables/new_crossencoder.tex
\begin{tabular}{lcccc}
\toprule
Corpus & Bits & Flip & Median sep.\ ratio \\
\midrule
SciFact  & W4 & 4.8\%  & 1.299 \\
SciFact  & W3 & 12.2\% & 0.454 \\
NFCorpus & W4 & 10.0\% & 0.783 \\
NFCorpus & W3 & 17.6\% & 0.308 \\
FiQA     & W4 & 9.8\%  & 0.591 \\
FiQA     & W3 & 27.0\% & 0.230 \\
\bottomrule
\end{tabular}

%% file: tables/new_quantizers.tex
\resizebox{\linewidth}{!}{\begin{tabular}{c|l|cccccccc}
\toprule
Bits & Quantizer & ViT-B/16 & ViT-L/16 & Qwen3-Emb-0.6B & GTE & BGE & E5 & Qwen3-Emb-4B & Qwen3-Emb-8B \\
\midrule
\multirow{6}{*}{\rotatebox[origin=c]{90}{W4}} 
& \rtn{} & 45.5\,{\scriptsize$\pm$2.5} & 42.1\,{\scriptsize$\pm$1.5} & 33.4\,{\scriptsize$\pm$4.1} & 20.0\,{\scriptsize$\pm$3.0} & 22.6\,{\scriptsize$\pm$3.1} & 29.5\,{\scriptsize$\pm$4.1} & 24.4\,{\scriptsize$\pm$3.2} & 21.3\,{\scriptsize$\pm$3.8} \\
& HQQ & 38.9\,{\scriptsize$\pm$2.6} & 33.5\,{\scriptsize$\pm$1.4} & 30.9\,{\scriptsize$\pm$3.4} & 17.3\,{\scriptsize$\pm$3.1} & 19.1\,{\scriptsize$\pm$2.6} & 22.0\,{\scriptsize$\pm$3.1} & 22.2\,{\scriptsize$\pm$3.1} & 18.2\,{\scriptsize$\pm$2.9} \\
& \gptq{} & 33.1\,{\scriptsize$\pm$1.3} & 29.9\,{\scriptsize$\pm$1.5} & 23.1\,{\scriptsize$\pm$3.0} & 15.6\,{\scriptsize$\pm$2.5} & 15.5\,{\scriptsize$\pm$2.6} & 19.3\,{\scriptsize$\pm$3.4} & 17.3\,{\scriptsize$\pm$2.4} & 15.7\,{\scriptsize$\pm$3.0} \\
& \gptq{} (act-order) & 30.6\,{\scriptsize$\pm$1.6} & 29.6\,{\scriptsize$\pm$1.3} & 22.5\,{\scriptsize$\pm$3.3} & 14.9\,{\scriptsize$\pm$2.7} & 14.4\,{\scriptsize$\pm$2.3} & 16.7\,{\scriptsize$\pm$2.7} & 18.4\,{\scriptsize$\pm$3.0} & 14.5\,{\scriptsize$\pm$3.0} \\
& \awq{} & 43.6\,{\scriptsize$\pm$2.3} & 41.4\,{\scriptsize$\pm$1.2} & 29.1\,{\scriptsize$\pm$3.5} & 16.8\,{\scriptsize$\pm$2.8} & 17.1\,{\scriptsize$\pm$2.4} & 18.5\,{\scriptsize$\pm$2.1} & 21.0\,{\scriptsize$\pm$3.2} & 17.0\,{\scriptsize$\pm$2.9} \\
& AdaRound & 33.8\,{\scriptsize$\pm$1.5} & 32.1\,{\scriptsize$\pm$1.6} & 28.7\,{\scriptsize$\pm$3.1} & 17.9\,{\scriptsize$\pm$2.7} & 18.9\,{\scriptsize$\pm$2.7} & 23.7\,{\scriptsize$\pm$3.4} & 20.1\,{\scriptsize$\pm$2.7} & 18.2\,{\scriptsize$\pm$3.0} \\
\midrule
\multirow{6}{*}{\rotatebox[origin=c]{90}{W3}} 
& \rtn{} & 80.9\,{\scriptsize$\pm$2.0} & 92.4\,{\scriptsize$\pm$0.9} & 86.0\,{\scriptsize$\pm$4.1} & 44.8\,{\scriptsize$\pm$4.8} & 54.3\,{\scriptsize$\pm$6.5} & 70.9\,{\scriptsize$\pm$9.4} & 75.2\,{\scriptsize$\pm$6.3} & 46.9\,{\scriptsize$\pm$6.4} \\
& HQQ & 67.2\,{\scriptsize$\pm$2.5} & 61.7\,{\scriptsize$\pm$1.4} & 59.7\,{\scriptsize$\pm$5.8} & 35.2\,{\scriptsize$\pm$4.1} & 43.3\,{\scriptsize$\pm$5.5} & 46.6\,{\scriptsize$\pm$6.8} & 44.5\,{\scriptsize$\pm$6.1} & 34.1\,{\scriptsize$\pm$5.8} \\
& \gptq{} & 61.5\,{\scriptsize$\pm$1.8} & 57.8\,{\scriptsize$\pm$2.5} & 54.7\,{\scriptsize$\pm$5.7} & 31.7\,{\scriptsize$\pm$4.7} & 36.5\,{\scriptsize$\pm$4.8} & 39.9\,{\scriptsize$\pm$5.8} & 48.4\,{\scriptsize$\pm$6.3} & 41.5\,{\scriptsize$\pm$6.6} \\
& \gptq{} (act-order) & 59.0\,{\scriptsize$\pm$2.2} & 55.4\,{\scriptsize$\pm$2.5} & 53.4\,{\scriptsize$\pm$6.1} & 31.4\,{\scriptsize$\pm$4.6} & 33.6\,{\scriptsize$\pm$4.6} & 38.9\,{\scriptsize$\pm$4.9} & 48.4\,{\scriptsize$\pm$6.3} & 41.0\,{\scriptsize$\pm$6.3} \\
& \awq{} & 78.3\,{\scriptsize$\pm$1.9} & 89.5\,{\scriptsize$\pm$1.4} & 69.1\,{\scriptsize$\pm$5.7} & 34.5\,{\scriptsize$\pm$3.8} & 38.5\,{\scriptsize$\pm$4.3} & 40.2\,{\scriptsize$\pm$5.5} & 47.6\,{\scriptsize$\pm$6.7} & 37.2\,{\scriptsize$\pm$5.8} \\
& AdaRound & 62.5\,{\scriptsize$\pm$2.2} & 59.6\,{\scriptsize$\pm$1.9} & 65.3\,{\scriptsize$\pm$7.4} & 36.7\,{\scriptsize$\pm$4.5} & 45.4\,{\scriptsize$\pm$6.4} & 52.6\,{\scriptsize$\pm$7.1} & 48.2\,{\scriptsize$\pm$5.1} & 38.7\,{\scriptsize$\pm$6.7} \\
\bottomrule
\end{tabular}}

%% file: tables/new_regime.tex
\begin{tabular}{llcccccccccc}
\toprule
& & \multicolumn{5}{c}{ViT-B/16} & \multicolumn{5}{c}{ViT-L/16} \\
\cmidrule(lr){3-7} \cmidrule(lr){8-12}
& & \multicolumn{3}{c}{Classification} & \multicolumn{2}{c}{Retrieval} & \multicolumn{3}{c}{Classification} & \multicolumn{2}{c}{Retrieval} \\
\cmidrule(lr){3-5}\cmidrule(lr){6-7} \cmidrule(lr){8-10}\cmidrule(lr){11-12}
Quantizer & Bits & Sep & Change & Acc & Sep & Change & Sep & Change & Acc & Sep & Change \\
\midrule
\rtn{} & W4 & 3.18 & 6.2\% & 81.0\% & 0.017 & 45.5\% & 5.63 & 2.9\% & 88.9\% & 0.019 & 42.1\% \\
HQQ & W4 & 4.61 & 5.7\% & 81.5\% & 0.019 & 38.9\% & 8.30 & 2.1\% & 89.2\% & 0.027 & 33.5\% \\
\gptq{} & W4 & 10.91 & 2.9\% & 82.0\% & 0.029 & 33.1\% & 14.05 & 1.4\% & 89.3\% & 0.044 & 29.9\% \\
\gptq{} (act-order) & W4 & 13.11 & 2.5\% & 82.0\% & 0.033 & 30.6\% & 15.76 & 1.3\% & 89.3\% & 0.047 & 29.6\% \\
\awq{} & W4 & 4.17 & 6.2\% & 81.3\% & 0.017 & 43.6\% & 5.42 & 3.0\% & 88.9\% & 0.019 & 41.4\% \\
AdaRound & W4 & 8.06 & 3.4\% & 81.8\% & 0.027 & 33.8\% & 9.35 & 1.7\% & 89.3\% & 0.031 & 32.1\% \\
\midrule
\rtn{} & W3 & 0.64 & 25.4\% & 69.6\% & 0.006 & 80.9\% & 0.51 & 42.1\% & 55.5\% & 0.005 & 92.4\% \\
HQQ & W3 & 1.19 & 12.1\% & 78.6\% & 0.009 & 67.2\% & 2.03 & 6.0\% & 87.5\% & 0.009 & 61.7\% \\
\gptq{} & W3 & 3.06 & 7.0\% & 81.0\% & 0.012 & 61.5\% & 3.01 & 3.8\% & 88.6\% & 0.017 & 57.8\% \\
\gptq{} (act-order) & W3 & 4.07 & 6.2\% & 81.2\% & 0.013 & 59.0\% & 3.39 & 3.4\% & 88.8\% & 0.018 & 55.4\% \\
\awq{} & W3 & 0.81 & 19.6\% & 72.9\% & 0.007 & 78.3\% & 0.58 & 33.8\% & 63.1\% & 0.006 & 89.5\% \\
AdaRound & W3 & 2.23 & 8.4\% & 80.4\% & 0.012 & 62.5\% & 2.48 & 4.8\% & 88.2\% & 0.013 & 59.6\% \\
\bottomrule
\end{tabular}

%% file: tables/new_alloc_absolute.tex
\begin{tabular}{l|ccc}
\toprule
Configuration & Flip $\downarrow$ & R@1 $\uparrow$ & Gold lost $\downarrow$ \\
\midrule
All W3 & 61.0\% & 35.1\% & 40.4\% \\
Reconstruction error & 43.1\% & 46.0\% & 19.9\% \\
\textbf{Gap sensitivity} & \textbf{33.3\%} & \textbf{48.5\%} & \textbf{13.8\%} \\
All W4 & 23.3\% & 50.5\% & 8.5\% \\
\bottomrule
\end{tabular}

%% file: tables/new_alloc_rag_gap.tex
\begin{tabular}{lccc}
\toprule
Quantizer & Capture & Changed & R$\to$W \\
\midrule
\rtn{} & 73.3\% & 770 & 201 \\
HQQ & 70.5\% & 654 & 128 \\
\gptq{} & 63.6\% & 542 & 86 \\
\gptq{} (AO) & 77.6\% & 532 & 91 \\
\awq{} & 66.6\% & 649 & 124 \\
AdaRound & 70.4\% & 629 & 113 \\
\bottomrule
\end{tabular}

%% file: tables/new_alloc_rag_mse.tex
\begin{tabular}{lccc}
\toprule
Quantizer & Capture & Changed & R$\to$W \\
\midrule
\rtn{} & 59.0\% & 895 & 244 \\
HQQ & 57.0\% & 690 & 153 \\
\gptq{} & 41.1\% & 600 & 114 \\
\gptq{} (AO) & 53.1\% & 595 & 112 \\
\awq{} & 51.2\% & 725 & 158 \\
AdaRound & 54.8\% & 641 & 122 \\
\bottomrule
\end{tabular}

%% file: tables/new_alloc_permodel.tex
\begin{tabular}{lr|rccc}
\toprule
Model & Reconstruction error & Gap sensitivity & Layers $n$ & Mean $\rho$ & Range across runs \\
\midrule
Qwen3-Emb-0.6B & 55.6\% & \textbf{68.3\%} & 196 & $-0.037$ & $[-0.15, +0.03]$ \\
Qwen3-Emb-4B & 40.9\% & \textbf{73.5\%} & 252 & $+0.118$ & $[+0.06, +0.18]$ \\
Qwen3-Emb-8B & 47.1\% & \textbf{73.3\%} & 252 & $+0.150$ & $[+0.11, +0.21]$ \\
BGE-large-en-v1.5 & 37.0\% & \textbf{72.7\%} & 145 & $-0.101$ & $[-0.16, -0.07]$ \\
E5-large-v2 & 57.8\% & \textbf{84.5\%} & 145 & $-0.278$ & $[-0.33, -0.22]$ \\
GTE-large & 40.7\% & \textbf{72.0\%} & 145 & $-0.101$ & $[-0.14, -0.07]$ \\
\bottomrule
\end{tabular}

%% file: tables/new_routing_sweep.tex
    \begin{tabular}{llcccccc}
    \toprule
    & & \multicolumn{2}{c}{10th pct.} & \multicolumn{2}{c}{25th pct.} & \multicolumn{2}{c}{50th pct.} \\
    \cmidrule(lr){3-4}\cmidrule(lr){5-6}\cmidrule(lr){7-8}
    Backbone & Score & Routed & Recovered & Routed & Recovered & Routed & Recovered \\
    \midrule
    ViT-B/16 (17) & Gap (ours) & 9.8\% & \textbf{62.3\%} & 24.9\% & \textbf{84.8\%} & 50.8\% & \textbf{96.6\%} \\
     & MSP & 10.1\% & 58.4\% & 25.0\% & 79.5\% & 49.6\% & 92.7\% \\
     & Entropy & 9.8\% & 48.3\% & 24.8\% & 69.0\% & 49.7\% & 83.6\% \\
    \addlinespace
    ViT-L/16 (16) & Gap (ours) & 10.2\% & \textbf{71.9\%} & 24.5\% & \textbf{92.5\%} & 50.5\% & \textbf{99.3\%} \\
     & MSP & 9.6\% & 68.6\% & 24.5\% & 89.6\% & 49.9\% & 98.2\% \\
     & Entropy & 9.5\% & 60.1\% & 24.3\% & 84.2\% & 49.6\% & 98.0\% \\
    \addlinespace
    Qwen3-Emb-0.6B (8) & Gap (ours) & 10.1\% & \textbf{75.7\%} & 25.0\% & \textbf{87.8\%} & 50.7\% & \textbf{97.9\%} \\
     & MSP & 10.1\% & 74.8\% & 25.4\% & 88.2\% & 50.0\% & 98.3\% \\
     & Entropy & 10.3\% & 73.0\% & 25.4\% & 88.1\% & 50.2\% & 98.3\% \\
    \bottomrule
    \end{tabular}

%% file: tables/new_granularity.tex
\begin{tabular}{lcccccc}
\toprule
& \multicolumn{3}{c}{Classification} & \multicolumn{3}{c}{Retrieval} \\
\cmidrule(lr){2-4}\cmidrule(lr){5-7}
Quantizer & Per channel  & Group 128  & Group 64 & Per channel  & Group 128  & Group 64 \\
\midrule
\rtn{} & 42.1\% & 18.1\% & 9.6\% & 69.9\% & 50.0\% & 45.2\% \\
HQQ & 23.1\% & 6.6\% & 5.5\% & 57.1\% & 37.1\% & 32.9\% \\
\gptq{} & 8.3\% & 4.5\% & 3.9\% & 48.0\% & 33.3\% & 30.1\% \\
\gptq{} (act-order) & 8.0\% & 4.1\% & 3.6\% & 46.9\% & 32.1\% & 28.9\% \\
\awq{} & 37.4\% & 14.4\% & 8.0\% & 63.2\% & 40.9\% & 36.1\% \\
AdaRound & 11.9\% & 5.1\% & 4.5\% & 56.8\% & 37.1\% & 33.6\% \\
\bottomrule
\end{tabular}

%% file: tables/new_rag.tex
\begin{tabular}{lcccccc}
\toprule
Condition & Exact match & Token F1 & Gold in context & Answers changed & Right$\to$wrong & Wrong$\to$right \\
\midrule
No passages & 15.1\% & 26.8\% & -- & -- & -- & -- \\
\fp{} retrieval & 42.7\% & 58.1\% & 65.0\% & -- & -- & -- \\
\ptq{} retrieval, \rtn{} & 41.2\% & 56.4\% & 61.8\% & 731 & 106 & 70 \\
\ptq{} retrieval, HQQ & 42.0\% & 57.2\% & 62.4\% & 677 & 101 & 83 \\
\ptq{} retrieval, \gptq{} & 41.7\% & 57.1\% & 63.9\% & 548 & 77 & 54 \\
\ptq{} retrieval, \gptq{} (AO) & 42.7\% & 58.0\% & 64.5\% & 589 & 65 & 64 \\
\ptq{} retrieval, \awq{} & 41.9\% & 57.4\% & 62.9\% & 675 & 90 & 70 \\
\ptq{} retrieval, AdaRound & 41.8\% & 57.0\% & 63.3\% & 631 & 82 & 60 \\
\bottomrule
\end{tabular}

%% file: tables/new_topk.tex
\begin{tabular}{l|cccc|cccc}
\toprule
& \multicolumn{4}{c}{Separation at the $k$-cut} & \multicolumn{4}{c}{Exact top-$k$ set kept} \\
\cmidrule(lr){2-5}\cmidrule(lr){6-9}
Backbone & $k{=}1$ & $k{=}2$ & $k{=}3$ & $k{=}5$ & $k{=}1$ & $k{=}2$ & $k{=}3$ & $k{=}5$ \\
\midrule
ViT-B/16 & 4.04 & 0.38 & 0.20 & 0.11 & 94\% & 76\% & 62\% & 45\% \\
ViT-L/16 & 5.36 & 0.46 & 0.23 & 0.12 & 97\% & 81\% & 66\% & 47\% \\
Qwen3-Emb & 2.42 & 0.19 & 0.11 & 0.08 & 97\% & 62\% & 42\% & 28\% \\
\bottomrule
\end{tabular}

%% file: tables/new_accrank.tex
\begin{tabular}{l|lrrc}
\toprule
Model & Configuration & $n$ & Spearman & $p$ \\
\midrule
ViT-B/16 & W4-channel & 19 & $+0.174$ & 0.477 \\
ViT-B/16 & W4-group\_128 & 19 & $+0.089$ & 0.716 \\
ViT-B/16 & W3-channel & 19 & $+0.247$ & 0.307 \\
ViT-B/16 & W3-group\_128 & 19 & $+0.214$ & 0.379 \\
ViT-L/16 & W4-channel & 19 & $+0.461$$^{*}$ & 0.047 \\
ViT-L/16 & W4-group\_128 & 19 & $+0.230$ & 0.344 \\
ViT-L/16 & W3-channel & 19 & $+0.655$$^{*}$ & 0.002 \\
ViT-L/16 & W3-group\_128 & 19 & $+0.496$$^{*}$ & 0.031 \\
\bottomrule
\end{tabular}

%% file: tables/new_harm.tex
\begin{tabular}{l|ccc|ccc}
\toprule
& \multicolumn{3}{c}{W4-group\_128} & \multicolumn{3}{c}{W3-group\_128} \\
\cmidrule(lr){2-4}\cmidrule(lr){5-7}
Separation $\gapk{2}/2\varepsilon$ & Queries & Flip & Harm & Queries & Flip & Harm \\
\midrule
$[0,\, 0.05)$ & 16,508 & 53.0\% & 6.4\% & 28,690 & 69.7\% & 11.2\% \\
$[0.05,\, 0.1)$ & 11,069 & 30.5\% & 3.9\% & 12,746 & 47.5\% & 11.4\% \\
$[0.1,\, 0.2)$ & 14,175 & 13.9\% & 2.5\% & 10,657 & 25.6\% & 8.2\% \\
$[0.2,\, 0.5)$ & 16,011 & 1.9\% & 0.4\% & 6,350 & 6.1\% & 3.4\% \\
$[0.5,\, 1)$ & 5,715 & 0.0\% & 0.0\% & 954 & 0.2\% & 0.2\% \\
$\ge 1$ & 2,097 & 0.0\% & 0.0\% & -- & -- & -- \\
\bottomrule
\end{tabular}

%% file: tables/new_corpus.tex
\begin{tabular}{llrrccrr}
\toprule
Backbone & Dataset & \multicolumn{2}{c}{Corpus size} & \multicolumn{2}{c}{Change rate} & \multicolumn{2}{c}{Separation} \\
\cmidrule(lr){3-4}\cmidrule(lr){5-6}\cmidrule(lr){7-8}
 & & Small & Large & Small & Large & Small & Large \\
\midrule
Qwen3-Emb & FiQA & 1{,}000 & 57{,}638 & 42.7\% & 41.9\% & 0.069 & 0.058 \\
Qwen3-Emb & NFCorpus & 1{,}000 & 3{,}633 & 36.1\% & 34.5\% & 0.067 & 0.062 \\
Qwen3-Emb & SCIDOCS & 1{,}000 & 25{,}657 & 31.8\% & 35.2\% & 0.087 & 0.051 \\
Qwen3-Emb & SciFact & 1{,}000 & 5{,}183 & 28.8\% & 22.0\%$^{*}$ & 0.091 & 0.132 \\
ViT-B/16 & CIFAR100 & 1{,}000 & 10{,}000 & 33.3\% & 46.0\%$^{*}$ & 0.048 & 0.013 \\
ViT-B/16 & Cars & 1{,}000 & 8{,}041 & 43.6\% & 52.2\%$^{*}$ & 0.028 & 0.015 \\
ViT-B/16 & Food101 & 1{,}000 & 25{,}250 & 31.4\% & 49.7\%$^{*}$ & 0.060 & 0.017 \\
ViT-B/16 & SUN397 & 1{,}000 & 19{,}850 & 22.7\% & 39.4\%$^{*}$ & 0.145 & 0.041 \\
ViT-B/16 & TinyImageNet & 1{,}000 & 10{,}000 & 25.1\% & 40.5\%$^{*}$ & 0.099 & 0.026 \\
\bottomrule
\end{tabular}

%% file: tables/new_baselines.tex
\begin{tabular}{lcccccc}
\toprule
& \multicolumn{2}{c}{$\alpha=0.10$} & \multicolumn{2}{c}{$\alpha=0.05$} & \multicolumn{2}{c}{$\alpha=0.01$} \\
\cmidrule(lr){2-3}\cmidrule(lr){4-5}\cmidrule(lr){6-7}
Score & Cover & Viol. & Cover & Viol. & Cover & Viol. \\
\midrule
Our stability check & 85.2\% & 1.47\% & 80.0\% & 0.72\% & 67.7\% & 0.10\% \\
Margin, generic calibration & 82.3\% & 0.94\% & 77.9\% & 0.54\% & 69.1\% & 0.15\% \\
Max-softmax-prob, generic calibration & 74.4\% & 1.29\% & 69.4\% & 0.81\% & 61.1\% & 0.33\% \\
\bottomrule
\end{tabular}

%% file: tables/new_certificate_tau.tex
\begin{tabular}{l|ccc}
\toprule
Setting & $\alpha = 0.10$ & $\alpha = 0.05$ & $\alpha = 0.01$ \\
\midrule
Classification (ViT-B/16) & 0.174 & 0.286 & 0.787 \\
Text retrieval (Qwen3) & 0.0854 & 0.0982 & 0.139 \\
\textsc{clip} ViT-B/32 & 0.0571 & 0.065 & 0.0779 \\
\textsc{clip} ViT-L/14 & 0.0454 & 0.0503 & 0.0629 \\
\bottomrule
\end{tabular}

%% file: tables/new_pflip.tex
\begin{tabular}{llccc}
\toprule
Group & $s$ & $|\cset|$ & Predicted & Measured \\
\midrule
Classification, ViT-B/16 & 0.92 & 1 & 12.2\% & 15.6\% \\
 &  & 2 & 19.9\% & 20.1\% \\
 &  & 3 & 24.5\% & 23.5\% \\
 &  & 4 & 30.0\% & 30.0\% \\
\addlinespace
Classification, ViT-L/16 & 0.87 & 1 & 11.5\% & 13.7\% \\
 &  & 2 & 18.4\% & 17.0\% \\
 &  & 3 & 23.1\% & 19.8\% \\
 &  & 4 & 25.7\% & 23.7\% \\
\addlinespace
Retrieval, ViT-B/16 & 0.12 & 1--50 & 26.8\% & 28.9\% \\
 &  & 51--120 & 38.9\% & 36.7\% \\
 &  & 181--199 & 48.6\% & 49.1\% \\
\addlinespace
Retrieval, Qwen3-Emb & 0.24 & 1--50 & 6.5\% & 4.1\% \\
 &  & 181--199 & 36.7\% & 36.8\% \\
\bottomrule
\end{tabular}

%% file: tables/new_clip_mechanism.tex
\begin{tabular}{l|l|ccccc}
\toprule
Encoder & Bits & Change & Separation & $\cset = \emptyset$ & From runner-up & Top-5 set kept \\
\midrule
ViT-B/32 & W4 & 47.0\% & 0.079 & 0.0\% & 33.3\% & 2.5\% \\
ViT-B/32 & W3 & 91.9\% & 0.037 & 0.0\% & 4.7\% & 0.0\% \\
ViT-L/14 & W4 & 24.6\% & 0.151 & 1.1\% & 49.3\% & 11.7\% \\
ViT-L/14 & W3 & 58.1\% & 0.062 & 0.0\% & 19.3\% & 0.5\% \\
\bottomrule
\end{tabular}

%% file: tables/new_clip_regime.tex
\begin{tabular}{llcccccc}
\toprule
& & \multicolumn{3}{c}{ViT-B/32} & \multicolumn{3}{c}{ViT-L/14} \\\cmidrule(lr){3-5}\cmidrule(lr){6-8} Quantizer & Bits & Sep & Change & R@1 & Sep & Change & R@1 \\
\midrule
\rtn{} & W4 & 0.079 & 47.0\% & 41.9\% & 0.151 & 24.6\% & 59.1\% \\
HQQ & W4 & 0.101 & 34.4\% & 46.5\% & 0.176 & 21.2\% & 59.1\% \\
\gptq{} & W4 & 0.165 & 21.9\% & 48.9\% & 0.204 & 16.0\% & 60.5\% \\
\gptq{} (act-order) & W4 & 0.181 & 21.3\% & 48.6\% & 0.278 & 16.6\% & 60.0\% \\
\awq{} & W4 & 0.110 & 31.6\% & 47.8\% & 0.170 & 22.4\% & 60.2\% \\
AdaRound & W4 & 0.125 & 28.1\% & 47.9\% & 0.197 & 17.4\% & 60.4\% \\
\midrule
\rtn{} & W3 & 0.037 & 91.9\% & 9.4\% & 0.062 & 58.1\% & 40.1\% \\
HQQ & W3 & 0.053 & 65.0\% & 32.4\% & 0.090 & 40.6\% & 52.9\% \\
\gptq{} & W3 & 0.069 & 52.4\% & 40.3\% & 0.114 & 33.1\% & 56.9\% \\
\gptq{} (act-order) & W3 & 0.076 & 47.8\% & 43.1\% & 0.116 & 33.5\% & 55.6\% \\
\awq{} & W3 & 0.047 & 74.1\% & 26.0\% & 0.074 & 42.7\% & 53.1\% \\
AdaRound & W3 & 0.062 & 56.1\% & 37.5\% & 0.095 & 38.1\% & 55.0\% \\
\bottomrule
\end{tabular}

%% file: tables/new_clip_corpus.tex
\begin{tabular}{llrrccrr}
\toprule
Encoder & Dataset & \multicolumn{2}{c}{Corpus size} & \multicolumn{2}{c}{Change rate} & \multicolumn{2}{c}{Separation} \\ \cmidrule(lr){3-4}\cmidrule(lr){5-6}\cmidrule(lr){7-8} & & Small & Large & Small & Large & Small & Large \\
\midrule
ViT-B/32 & Flickr30k & 500 & 4{,}000 & 47.4\% & 47.0\% & 0.084 & 0.079 \\
ViT-L/14 & Flickr30k & 500 & 4{,}000 & 34.2\% & 24.6\%$^{*}$ & 0.127 & 0.151 \\
\bottomrule
\end{tabular}

%% file: tables/new_clip_alloc_main.tex
\begin{tabular}{l|lrrr}
\toprule
Quantizer & Allocation & Flip & R@1 & Gold kept \\
\midrule
\rtn{} & Reconstruction error & 37.8\% & 47.3\% & 43.4\% \\
 & Gap sensitivity & \textbf{60.9\%} & \textbf{74.3\%} & \textbf{68.8\%} \\
\midrule
\gptq{} & Reconstruction error & 22.4\% & 23.4\% & 23.6\% \\
 & Gap sensitivity & \textbf{37.9\%} & \textbf{32.8\%} & \textbf{39.5\%} \\
\midrule
\awq{} & Reconstruction error & 15.6\% & 21.3\% & 18.6\% \\
 & Gap sensitivity & \textbf{44.8\%} & \textbf{67.1\%} & \textbf{55.0\%} \\
\bottomrule
\end{tabular}

%% file: tables/new_clip_alloc_absolute.tex
\begin{tabular}{l|ccc}
\toprule
Configuration & Flip $\downarrow$ & R@1 $\uparrow$ & Gold lost $\downarrow$ \\
\midrule
All W3 & 74.8\% & 24.9\% & 65.6\% \\
Reconstruction error & 60.2\% & 37.0\% & 45.6\% \\
\textbf{Gap sensitivity} & \textbf{51.2\%} & \textbf{43.9\%} & \textbf{33.9\%} \\
All W4 & 36.1\% & 50.5\% & 19.6\% \\
\bottomrule
\end{tabular}

%% file: tables/new_clip_alloc_permodel.tex
\begin{tabular}{l|rrccc}
\toprule
Model & Reconstruction error & Gap sensitivity & Layers $n$ & Mean $\rho$ & Range across runs \\
\midrule
ViT-B/32 & 25.8\% & \textbf{57.3\%} & 72 & $+0.116$ & $[+0.09, +0.12]$ \\
ViT-L/14 & 53.9\% & \textbf{65.6\%} & 108 & $+0.342$ & $[+0.33, +0.37]$ \\
\bottomrule
\end{tabular}

%% file: tables/new_alloc_budget.tex
\begin{tabular}{ll|l|rrr}
\toprule
System & Budget & Allocation & Capture: flip & Capture: R@1 & Capture: gold \\
\midrule
Text & 3.25 bits & Reconstruction error & 28.4\% & 45.5\% & 42.5\% \\
 &  & Gap sensitivity & \textbf{46.4\%} & \textbf{68.3\%} & \textbf{62.1\%} \\
 & 3.5 bits & Reconstruction error & 47.5\% & 70.4\% & 64.3\% \\
 &  & Gap sensitivity & \textbf{73.4\%} & \textbf{86.9\%} & \textbf{83.2\%} \\
 & 3.75 bits & Reconstruction error & 63.6\% & 79.3\% & 76.5\% \\
 &  & Gap sensitivity & \textbf{89.8\%} & \textbf{97.5\%} & \textbf{94.4\%} \\
\midrule
\textsc{clip} & 3.25 bits & Reconstruction error & 15.7\% & 22.4\% & 18.9\% \\
 &  & Gap sensitivity & \textbf{32.4\%} & \textbf{41.1\%} & \textbf{37.9\%} \\
 & 3.5 bits & Reconstruction error & 37.8\% & 47.3\% & 43.4\% \\
 &  & Gap sensitivity & \textbf{60.9\%} & \textbf{74.3\%} & \textbf{68.8\%} \\
 & 3.75 bits & Reconstruction error & 69.0\% & 79.9\% & 74.7\% \\
 &  & Gap sensitivity & \textbf{83.5\%} & \textbf{93.4\%} & \textbf{87.9\%} \\
\bottomrule
\end{tabular}

%% file: tables/new_alloc_seed.tex
\begin{tabular}{l|l|rr}
\toprule
System & Model & Seed pairs & Layer agreement \\
\midrule
Text & Qwen3-Emb-0.6B & 12 & 88.8\% \\
 & Qwen3-Emb-4B & 12 & 90.0\% \\
 & Qwen3-Emb-8B & 12 & 93.0\% \\
 & BGE-large-en-v1.5 & 12 & 93.0\% \\
 & E5-large-v2 & 12 & 90.7\% \\
 & GTE-large & 12 & 92.9\% \\
\midrule
 & All & 72 & \textbf{91.4\%} \\
\midrule
\textsc{clip} & ViT-B/32 & 3 & 90.7\% \\
 & ViT-L/14 & 3 & 87.7\% \\
\midrule
 & All &  & \textbf{88.9\%} \\
\bottomrule
\end{tabular}

%% file: tables/new_alloc_calib.tex
\resizebox{\linewidth}{!}{\begin{tabular}{lrcccccccc}
\toprule
& & & \multicolumn{7}{c}{Capture by criterion} \\
\cmidrule(lr){4-10}
System & Queries & Agreement & Gap & Act.\ norm & Recon. & SQNR & Depth & Act.-wt. & Salience \\
\midrule
Text & 8 & 80.6\% & \textbf{66.6\%} & 51.4\% & 45.6\% & 45.8\% & 46.4\% & 42.4\% & 41.1\% \\
 & 32 & 89.2\% & \textbf{71.1\%} & 54.7\% & 45.7\% & 45.7\% & 46.5\% & 43.7\% & 40.9\% \\
 & 64 & 92.9\% & \textbf{72.4\%} & 52.6\% & 45.7\% & 45.7\% & 46.4\% & 43.1\% & 40.5\% \\
 & 128 & --- & \textbf{73.8\%} & 52.5\% & 45.6\% & 45.6\% & 46.3\% & 42.8\% & 41.0\% \\
 & 256 & 95.2\% & \textbf{73.2\%} & 52.7\% & 45.7\% & 45.5\% & 46.6\% & 42.9\% & 40.7\% \\
\midrule
\textsc{clip} ViT-B/32 & 8 & 86.1\% & \textbf{52.9\%} & 36.9\% & 26.2\% & 36.7\% & 35.8\% & 37.1\% & 29.7\% \\
 & 32 & 91.7\% & \textbf{53.4\%} & 36.9\% & 26.0\% & 36.3\% & 35.5\% & 37.2\% & 29.6\% \\
 & 64 & 94.4\% & \textbf{57.7\%} & 36.6\% & 25.8\% & 36.2\% & 35.2\% & 36.9\% & 29.5\% \\
 & 128 & --- & \textbf{57.3\%} & 36.0\% & 25.8\% & 35.9\% & 35.2\% & 36.3\% & 29.0\% \\
 & 256 & 94.4\% & \textbf{54.4\%} & 35.3\% & 25.8\% & 35.8\% & 34.7\% & 35.4\% & 27.9\% \\
\midrule
\textsc{clip} ViT-L/14 & 8 & 77.8\% & \textbf{64.0\%} & 42.9\% & 54.1\% & 44.1\% & 40.5\% & 41.7\% & 46.1\% \\
 & 32 & 88.0\% & \textbf{60.3\%} & 42.5\% & 54.1\% & 44.2\% & 40.8\% & 41.4\% & 45.9\% \\
 & 64 & 89.8\% & \textbf{65.3\%} & 41.9\% & 53.8\% & 43.6\% & 40.3\% & 41.2\% & 45.9\% \\
 & 128 & --- & \textbf{65.6\%} & 41.6\% & 53.9\% & 44.1\% & 40.8\% & 41.3\% & 46.5\% \\
 & 256 & 96.3\% & \textbf{65.4\%} & 41.3\% & 53.6\% & 43.1\% & 41.0\% & 41.5\% & 46.4\% \\
\bottomrule
\end{tabular}}

%% file: tables/new_activation.tex
\begin{tabular}{llccc}
\toprule
Family & & W4 & W4A8 & W8A8 \\
\midrule
Classification ($n=53$) & Top-1 change & 4.5\% & 8.7\% & 6.5\% \\
 & Median separation & 3.63 & 1.80 & 3.04 \\
 & Above threshold & 92\% & 75\% & 91\% \\
\midrule
Retrieval ($n=112$) & Top-1 change & 29.8\% & 60.1\% & 52.6\% \\
 & Median separation & 0.093 & 0.031 & 0.035 \\
 & Above threshold & 0\% & 0\% & 0\% \\
\bottomrule
\end{tabular}

%% file: tables/new_activation_quantizers.tex
\begin{tabular}{lcccccccccccc}
\toprule
& \multicolumn{6}{c}{W4} & \multicolumn{6}{c}{W8} \\
\cmidrule(lr){2-7}\cmidrule(lr){8-13}
& \multicolumn{2}{c}{Classification} & \multicolumn{2}{c}{Retrieval} & & & \multicolumn{2}{c}{Classification} & \multicolumn{2}{c}{Retrieval} & & \\
\cmidrule(lr){2-3}\cmidrule(lr){4-5}\cmidrule(lr){8-9}\cmidrule(lr){10-11}
Quantizer & W4 & W4A8 & W4 & W4A8 & Ratio & Spearman & W8 & W8A8 & W8 & W8A8 & Ratio & Spearman \\
\midrule
\rtn{} & 4.5\% & 8.7\% & 29.8\% & 60.1\% & 6.9$\times$ & $-0.90$ & 0.3\% & 6.5\% & 2.5\% & 52.6\% & 8.1$\times$ & $-0.93$ \\
HQQ & 3.9\% & 8.2\% & 24.9\% & 57.4\% & 7.0$\times$ & $-0.91$ & 0.2\% & 6.5\% & 2.3\% & 52.5\% & 8.0$\times$ & $-0.93$ \\
\gptq{} & 2.4\% & 7.4\% & 20.9\% & 57.1\% & 7.7$\times$ & $-0.92$ & 0.1\% & 6.4\% & 1.9\% & 52.6\% & 8.2$\times$ & $-0.93$ \\
\gptq{} (AO) & 2.2\% & 7.2\% & 19.8\% & 56.5\% & 7.8$\times$ & $-0.93$ & 0.1\% & 6.5\% & 1.8\% & 52.9\% & 8.1$\times$ & $-0.93$ \\
\awq{} & 4.4\% & 8.3\% & 25.6\% & 57.6\% & 7.0$\times$ & $-0.93$ & 0.2\% & 6.5\% & 2.3\% & 52.5\% & 8.1$\times$ & $-0.94$ \\
AdaRound & 2.8\% & 7.6\% & 23.8\% & 58.1\% & 7.6$\times$ & $-0.90$ & 0.2\% & 6.5\% & 2.2\% & 52.4\% & 8.1$\times$ & $-0.94$ \\
\bottomrule
\end{tabular}

%% file: tables/new_alloc_activation_quantizers.tex
\begin{tabular}{lcccccc}
\toprule
& \multicolumn{2}{c}{Gap sensitivity} & \multicolumn{2}{c}{Reconstruction error} & \multicolumn{2}{c}{Random split} \\
\cmidrule(lr){2-3}\cmidrule(lr){4-5}\cmidrule(lr){6-7}
Quantizer & W & W+A8 & W & W+A8 & W & W+A8 \\
\midrule
\rtn{} & \textbf{73.4\%} & \textbf{67.0\%} & 47.5\% & 37.0\% & 49.3\% & 37.7\% \\
HQQ & \textbf{72.8\%} & \textbf{75.9\%} & 43.8\% & 46.6\% & 45.0\% & 53.1\% \\
\gptq{} & \textbf{60.2\%} & \textbf{59.9\%} & 44.2\% & 52.6\% & 45.8\% & 49.8\% \\
\gptq{} (AO) & \textbf{63.2\%} & \textbf{59.8\%} & 43.9\% & 44.6\% & 46.9\% & 46.8\% \\
\awq{} & \textbf{60.7\%} & \textbf{63.5\%} & 46.6\% & 43.2\% & 49.8\% & 45.3\% \\
AdaRound & \textbf{69.4\%} & \textbf{62.7\%} & 39.8\% & 32.8\% & 41.6\% & 37.7\% \\
\bottomrule
\end{tabular}